\pdfoutput=1
\documentclass[11pt]{article}

\usepackage[preprint]{acl}
\usepackage{times}
\usepackage{latexsym}
\usepackage{dsfont}
\usepackage{comment}
\usepackage[group-separator={,}]{siunitx}
\usepackage[T1]{fontenc}

\usepackage[utf8]{inputenc}
\usepackage{adjustbox}
\usepackage{microtype}

\usepackage{inconsolata}
\usepackage{graphicx}
\usepackage{enumitem}
\usepackage{float} 
\usepackage{amsmath}
\usepackage{amssymb}
\usepackage{subcaption}
\usepackage{float}
\usepackage{authblk}
\usepackage{multirow}

\restylefloat{table}

\title{SciQAG: A Framework for Auto-Generated Science Question Answering Dataset with Fine-grained Evaluation}

\author[1,2]{Yuwei Wan}
\author[1]{Yixuan Liu}
\author[3]{Aswathy Ajith}
\author[4,5]{Clara Grazian}
\author[6]{Bram Hoex}
\author[7]{\authorcr \textbf{Wenjie Zhang}}
\author[2,*]{\textbf{Chunyu Kit}}
\author[1,6,*]{\textbf{Tong Xie}}
\author[3,*]{\textbf{Ian Foster}}

\affil[1]{GreenDynamics Pty. Ltd, Kensington, NSW, Australia}
\affil[2]{Department of Linguistics and Translation, City University of Hong Kong, Hong Kong, China}
\affil[3]{Department of Computer Science, University of Chicago, Chicago, IL, USA}
\affil[4]{School of Mathematics and Statistics, University of Sydney, Camperdown, NSW, Australia}
\affil[5]{DARE ARC Training Centre in Data Analytics for Resources and Environments, Australia}
\affil[6]{School of Photovoltaic and Renewable Energy Engineering, University of New South Wales, Kensington, NSW, Australia}
\affil[7]{School of Computer Science and Engineering, University of New South Wales, Kensington, NSW, Australia}

\affil[*]{Corresponding author: tong.xie@unsw.edu.au, foster@cs.uchicago.edu, ctckit@cityu.edu.hk}

\begin{document}
\maketitle
\begin{abstract}

We introduce SciQAG, a novel framework for automatically generating high-quality science question-answer pairs from a large corpus of scientific literature based on large language models (LLMs). SciQAG consists of a QA generator and a QA evaluator, which work together to extract diverse and research-level questions and answers from scientific papers. Utilizing this framework, we construct a large-scale, high-quality, open-ended science QA dataset containing 188,042 QA pairs extracted from 22,743 scientific papers across 24 scientific domains. We also introduce SciQAG-24D, a new benchmark task designed to evaluate the science question-answering ability of LLMs. Extensive experiments demonstrate that fine-tuning LLMs on the SciQAG dataset significantly improves their performance on both open-ended question answering and scientific tasks. To foster research and collaboration, we make the datasets, models, and evaluation codes publicly available, contributing to the advancement of science question answering and developing more interpretable and reasoning-capable AI systems.

\end{abstract}

\section{Introduction}

Question-answering (QA) datasets have been instrumental in advancing natural language processing (NLP) research by enabling the development of models that can understand and reason about textual information.
These datasets encompass various domains and question types, such as reading comprehension \cite{hermann2015teaching}, open-domain QA \cite{chen2017reading, longpre2021mkqa}, and visual QA \cite{schwenk2022okvqa, masry2022chartqa}. They serve as benchmarks for evaluating the performance of QA systems and drive the creation of more sophisticated models \cite{chen2017reading, rajpurkar-etal-2016-squad}. Among these, science QA datasets such as SciQ \cite{welbl-etal-2017-crowdsourcing} and SciencQA \cite{saikh_scienceqa_2022} have garnered significant attention due to their potential to assess models' ability to comprehend and apply scientific knowledge \cite{clark2018think, mihaylov-etal-2018-suit}. The importance of these datasets lies not only in their capacity to test models' understanding of scientific concepts but also in their ability to evaluate the models' reasoning capabilities when dealing with complex relationships and drawing conclusions based on experimental evidence.

\begin{figure}
  \includegraphics[width=\columnwidth]{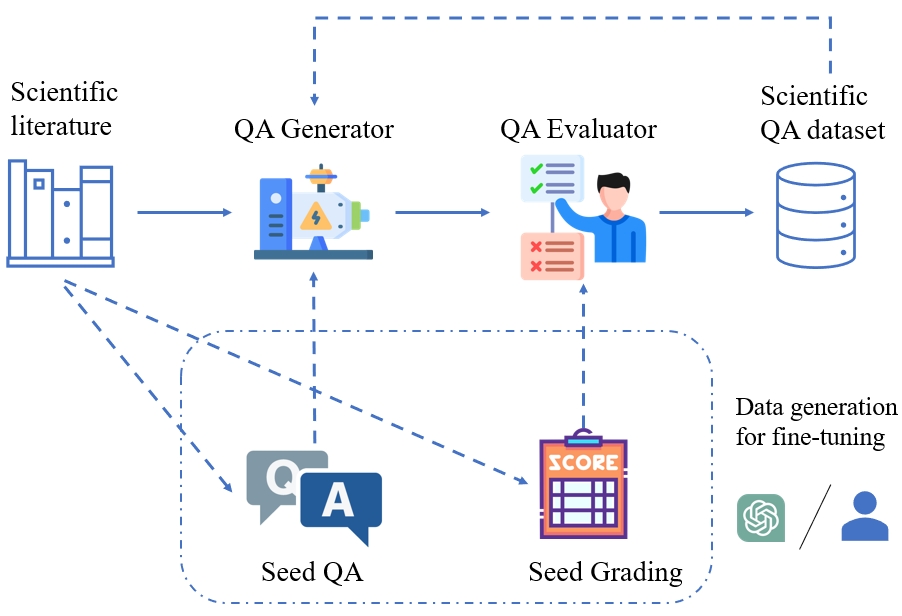}
  \caption{The SciQAG framework for generating science QA from the scientific literature. The dashed line represents optional fine-tuning.}
  \label{fig:1}
\end{figure}

The advent of large language models (LLMs) like GPT-4 \cite{openai2023gpt4}, LLaMA-3 \cite{touvron2023llama}, PaLM \cite{chowdhery2022palm}, marks a significant milestone in the pursuit of general artificial intelligence. These models have proven instrumental in various aspects, including generation, annotation, and paraphrasing \cite{yoo-etal-2021-gpt3mix-leveraging, xie2024creation,sahu-etal-2022-data, xie2023large, wang-etal-2023-self-instruct, tang2023metrics}. Besides general applications,this trend has permeated the natural sciences, giving rise to domain-specific models like BioBERT \cite{lee2020biobert}, Med-PALM \cite{singhal2022large}, and Galactica \cite{taylor2022galactica}. While these models may not always necessitate pre-training from scratch, it often involves fine-tuning on domain-specific data. QA pairs are essential for fine-tuning and evaluating LLMs \cite{kamalloo-etal-2023-evaluating, rasool2024evaluating}, particularly when domain-specific question-answering capabilities are desired. For example, \cite{zhang2024chemllm} fine-tuning data comprises 7M instruction tuning QA pairs, with the objective of instruction fine-tuning often being causal language modeling in GPT-based scientific LLMs \cite{zhang2024scientific}. As such, the development of more comprehensive science QA datasets and related evaluation methods is a practical necessity for advancing scientific LLMs.

Despite the progress made with existing science QA datasets, several challenges persist. First, these datasets often lack variety in question forms, with most being multiple-choice questions (MCQ). Although the performance of MCQs is established by exact match and conclusions obtained from such evaluation tend to be more robust \cite{subramanian2024mqalm}, they inadequately test the language and logical reasoning capabilities of LLMs, necessitating more challenging QA benchmarks. Second, the content of these datasets is primarily sourced from textbooks, resulting in an insufficient difficulty level to assess the high-level knowledge retention of LLMs in real-world academic research or industrial production settings. Third, the creation of these benchmarks often relies on crowd-sourcing or expert-driven approaches, which can be costly \cite{lee2023liquid, xie2023tokenizer} due to the need for specialized skills \cite{xie2023darwin, ye2024construction}. Addressing these challenges is crucial for developing robust and comprehensive QA datasets that can more accurately evaluate the capabilities of scientific LLMs. 

To tackle these challenges, we focus on the task of open-ended, closed-book science QA in this paper. Open-ended questions do not have predetermined or fixed answers, while closed-book indicates that no external knowledge or context is provided with the question  \cite{roberts2020much}. We introduce Automatic Generation of Science Question Answering (SciQAG), a framework for automatically generating QA pairs from scientific articles (see \autoref{fig:1}). SciQAG comprises two main components: a QA generator and a QA evaluator. The QA generator leverages LLMs to extract relevant information from scientific papers and generate diverse question-answer pairs while the QA evaluator assesses or filters the generated QA pairs. By applying SciQAG to a large corpus of scientific papers, we constructed a high-quality science QA dataset, SciQAG-24D, consisting of 188,042 QA pairs. We compared the zero-shot performance of several LLMs on the SciQAG-24D test set and explored fine-tuning impacts of the training set on various datasets, including unseen QA pairs, SciQ \cite{welbl-etal-2017-crowdsourcing}, and scientific tasks. The results demonstrate that incorporating science QA data during fine-tuning significantly enhances the model's ability to understand and apply scientific knowledge, improving performance on various scientific tasks. These findings underscore the value of the SciQAG-24D dataset and highlight the potential of scientific LLMs in advancing scientific research and problem-solving.

\section{Related Work}
\subsection{Science QA benchmarks}

\begin{table*}[ht!]
\centering
\begin{adjustbox}{width=0.9\textwidth}
\begin{tabular}{c|c|c|c|c}
\hline
\textbf{Dataset} & \textbf{Size} & \textbf{Method} & \textbf{Q-type} & \textbf{Difficulty} \\
\hline
SciQ \cite{welbl-etal-2017-crowdsourcing} & 13,679 & Manual & MCQ & Element - college \\
ScienceQA \cite{saikh_scienceqa_2022} & 21,208 & Manual & MCQ & Element - high \\
OpenBookQA \cite{mihaylov-etal-2018-suit} & 5,957 & Manual & MCQ & Element \\
ARC \cite{clark2018think} & 7,787 & Manual & MCQ & Element \\
MMLU \cite{hendrycks2020measuring} & 15,908 & Manual & MCQ & Element - profession \\
SciQA \cite{auer2023sciqa} & 2,565 & Automated & Open-ended & Profession \\
\hline
\end{tabular}
\end{adjustbox}
\caption{Overview of several popular science QA benchmark datasets, including their size, construction method, question type, and difficulty level.}
\label{table:1}
\end{table*}

Science QA benchmarks, which typically consist of questions and answers from scientific literature, textbooks, or domain experts, play a crucial role in evaluating the performance of QA systems in scientific domains. \autoref{table:1} provides a comparison of several popular science QA benchmarks, highlighting the prevalence of manually collected elementary-level multiple-choice questions Notable examples includes the work of \citet{welbl-etal-2017-crowdsourcing}, who proposed a novel approach to assist crowd workers in generating MCQs with answers from relevant passages. \citet{lu2022learn} compiled a multi-modal multiple-choice QA dataset from elementary and high school science curricula, emphasizing the evaluation of model reasoning skills. In the biomedical domain, \citet{jin-etal-2019-pubmedqa} produced a QA dataset based on PubMed article abstracts, focusing on reasoning beyond factoid answer extraction, but limited to \textit{yes}, \textit{no}, or \textit{maybe} answers.However, most previous open-ended long-form QA datasets on scientific literature are limited to domains like AI \cite{lee2023qasa, dasigi-etal-2021-dataset} or biomedical \cite{OTEGI2022108072, jin2020disease},leaving room for the development of more comprehensive and diverse science QA benchmarks.

\subsection{Automatic QA generation}
Automatic question generation has been explored in various contexts, with a primary focus on educational applications. Early methods utilized syntactic transformation templates \cite{mitkov2003computer, heilman2010good}to transform text into questions. \citet{pandya2021question} established a taxonomy for human-generated QA datasets and evaluation strategies for comparing model-generated answers with reference answers. However, current automatic question-generation methods typically rely on predefined answers  \cite{mulla_automatic_2023}. For example, \citet{virani2023automatic} introduced a versatile Question Answer Generation (QAG) system, adaptable to various question types and domains, but it still requires predefined answers for question generation. Recently, \citet{auer2023sciqa} used a rule-based method to automatically build SciQA for scholarly knowledge; but this approach relies on an existing knowledge graph. In contrast, the approach presented in this paper aims to generate high-quality open-ended QA pairs from scientific texts without relying on manually curated answers, thus addressing a gap in current literature. 

\subsection{Evaluation metrics for open-ended QA}
Evaluating the quality of long-form generated text, such as open-ended QA pairs, is a challenging task due to the subjective nature of the assessment  \cite{das2021automatic}. Traditional evaluation metrics, including BLEU \cite{papineni-etal-2002-bleu}, ROUGE \cite{lin-2004-rouge}, METEOR \cite{banerjee-lavie-2005-meteor}, and F1 \cite{rajpurkar-etal-2016-squad}, rely on \textit{n}-gram between the generated text and a reference text. More recent methods, such as BERTScore \cite{Zhang2020BERTScore}, sentence mover similarity \cite{clark-etal-2019-sentence}, and MAUVE \cite{pillutla2021mauve} eliminate the need for \textit{n}-gram similarities, but still require reference texts \cite{chen-etal-2019-evaluating}. \citet{min-etal-2023-factscore} propose an alternative approach that evaluates the factuality of long-form generated text by breaking it down into atomic facts and computing the percentage of atomic facts supported by a reliable knowledge source. The application of large language models (LLMs) to evaluate the quality of generated science QA datasets \cite{song-etal-2023-honeybee} has also shown promise. Building upon these advancements, this paper adopts a multi-perspective approach using LLMs to evaluate the quality of open-ended QA pairs, contributing to the development of more comprehensive evaluation metrics for science QA.

\section{SciQAG framework}
We propose a cost-effective framework to generating closed-book QA pairs focusing on scientific knowledge from the literature at scale,
called SciQAG. The main idea is to train a generator to convert full-text scientific papers into QA pairs and use an evaluator to filter out those that do not meet quality standards. 

\subsection{QA generator}
The task is defined as follows:
given seed input texts $T$, for each input text $t$, the generator should firstly generate 15 keywords $k$ that capture the most important terms and concepts in the text, then generate a set $S=\{(q_i, a_i)\}^n_{i=1}$ focusing on the generated keywords $k$, where $\forall i \in \{1, ..., 10\}$, $q_i$ is the question and $a_i$ is the answer to $q_i$. To generate $S$, one should learn a generator $G(S|T;\theta)$ with $\theta$ the model parameters. Thus, given a new input text $\hat{t}$, following $G(S|T;\theta)$, one can directly generate a $\hat{S}$ consisting of QA pairs (by firstly generating 15 keywords to guide the QA generation). 

We designed an effective prompt for LLM to generate QA pairs with detailed information extracted from scientific papers. Since our final generated QA dataset is intended to be closed-book, implying that models trained on our dataset will not have access to background knowledge during either training or inference, we also give guidelines to ensure that the generated QA do not depend on or reference the content in the original paper (see \autoref{sec:Appendix C} for prompts).  

For the dataset generation in this study, we chose to fine-tune an open-source LLM as the generator to balance performance and cost (see \autoref{sec:using}). However, users of this framework can select any open-source or closed-source LLM of their preference for the generator, either using fine-tuning or prompting.

\subsection{QA Evaluator}
Inspired by the evaluation metrics introduced in HoneyBee \cite{song-etal-2023-honeybee}, we develop a comprehensive evaluation metric, RACAR, which consists of five dimensions: \textbf{Relevance}, \textbf{Agnosticism}, \textbf{Completeness}, \textbf{Accuracy}, and \textbf{Reasonableness}. 

\begin{itemize}
\setlength{\itemsep}{0.1em}
\setlength{\parskip}{0.1em}
    \item \textbf{Relevance} quantifies \textit{how relevant} generated QA pairs are to the information presented in the source article. The generated questions are required to ask about the facts and knowledge provided in the article.
    \item \textbf{Agnosticism}: We require that the generated questions must not refer to demonstrative content from the original text, like figures or tables in the provided dataset. This dimension of the metric measures the degree of context-independence of the generated questions.
    \item \textbf{Completeness} assesses whether the answer comprehensively covers all relevant aspects of the question and effectively utilizes the details in the paper.
    \item \textbf{Accuracy} measures how well the generated answer corresponds to the relevant facts or information presented in the given paper. Any claims or statements in the answer should be supported by evidence from the paper.
    \item \textbf{Reasonableness} pertains to the internal logic within the generated answer. It involves assessing whether the answer is logically coherent without contradictions. 
\end{itemize}

\begin{figure}
  \includegraphics[width=\columnwidth]{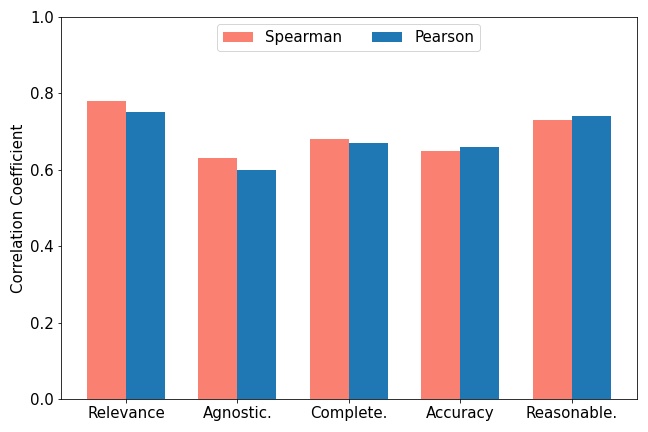}
  \caption{Spearman and Pearson correlations between GPT-4 assigned scores and expert-annotated scores.}
  \label{fig:2}
\end{figure}

To measure alignment between GPT-4 and human evaluation, we engaged two domain experts to manually evaluate the 10 QA pairs for 10 articles (a total of 100 pairs) with the same RACAR metric. \autoref{fig:2} shows the averaged spearman and pearson correlations between GPT-4 assigned scores and expert-annoted scores. 

Our evaluator serves two purposes: 1) evaluating the quality of the generated QA pairs, and 2) acting as a filter to discard low-quality QA pairs after establishing a standard. It is noteworthy that this standard is customizable; it can be a score threshold or a selection of the top k QA pairs based on their overall ranking. In this study, we directly use GPT-4 as QA evaluator to evaluate the generated QA pairs (see \autoref{sec:Appendix B} for prompts) along these dimensions on a scale of 1--5, with 1 being unacceptable and 5 fully acceptable. Users of this framework can select any open-source or cosed-source LLM for evaluator, either using fine-tuning or prompting.  
 
\begin{figure*}
\centering
  \includegraphics[width=0.85\linewidth]{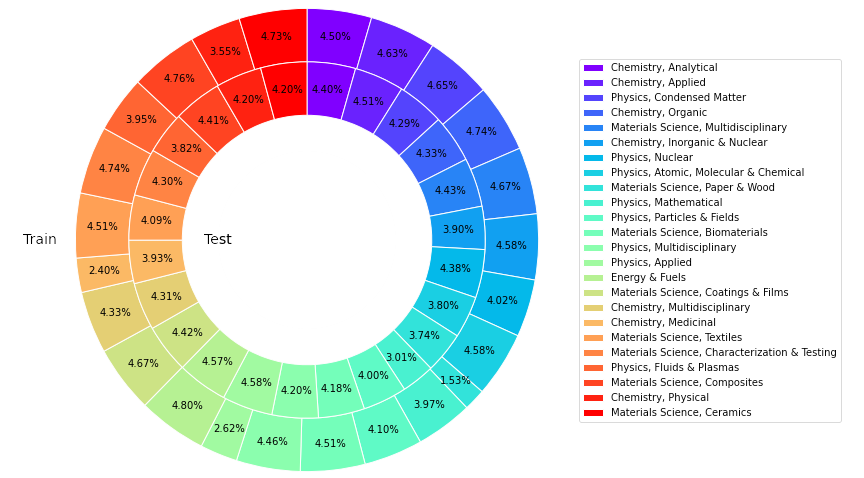}
  \caption{Proportion of papers of different categories in the training and testing of the SciQAG-24D dataset.}
\label{fig:ratio}
\end{figure*}
 
\section{SciQAG-24D Dataset}
\subsection{Paper collection}
From the Web of Science (WoS) Core Collection database, we obtained scientific papers that met the following criteria: 1) categorized in physical science disciplines (materials science, chemistry, physics, energy); 2) available in full-text HTML/XML format from publishers like Elsevier, Springer Nature, and Royal Society of Chemistry; and 3) written in the English language (see \autoref{sec:Appendix A} for details). We selected 24 WoS categories and for each WoS category, we selected most highly cited papers and employed the SciCrawler paper downloading tool \cite{xie2023darwin} to retrieve their content in plain text format. This dataset of \num{22743} papers aims to serve as a credible, rich, balanced, and representative source of scientific knowledge. The number of papers selected for each category can be found in \autoref{table:2} in \autoref{sec: statistics}.

\subsection{Using the SciQAG Framework}
\label{sec:using}
To fine-tune an open-source LLM as QA generator, we first randomly selected 426 papers from the paper collection as input to produce 4260 seed QA pairs by prompting GPT-4 (see \autoref{sec:Appendix B}). Then, we fine-tuned \texttt{vicuna-7b-v1.5-16k} model \cite{zheng2023judging} on seed data. The data employs the instruction schema \cite{wang2022self} composed of three elements: <instruction>, <input>, and <output>. The seed QA generation prompt was converted into the <instruction>. The seed paper filled the <input> field, and the <output> were the generated seed QA pairs. We concatenated the instruction and instance input as a prompt and train the model to generate the instance output in a standard supervised way. It is instructive to compare the SciQAG science QA generator with the self-questioning prompting approach of \citet{fang2024molinstructions}, which as we show in \autoref{sec:Appendix E} can also be used to generate factual QA pairs from text.

Using the trained QA generator, we performed inference on the remaining papers, generating a total of 227,430 QA pairs (including seed QA pairs). We sampled 50 papers from each category (a total of 1,200 papers) and their generated QA pairs, using GPT-4 to compute the RACAR score for each QA pair. QA pairs with any score lower than 3 were filtered out to form a test set. For the remaining QA pairs, a simple rule-based approach was used to remove all pairs containing "this paper" or "this study" to form a training set.

\subsection{Statistics of SciQAG-24D}
\textbf{Overview}. 
Using SciQAG framework, we established our open-ended science QA benchmark dataset, SciQAG-24D, which covers 24 scientific domains. \autoref{fig:ratio} shows the proportion of papers of different categories in the training and testing. And  \autoref{table:2} in \autoref{sec: statistics} lists the categories, the train-test split, the number of papers and generated QAs. The dataset consists of a total of 22,743 original papers, with 21,543 papers used for training set and 1,200 papers for test set. After filtering, the training set includes 21,529 papers and 179,511 QA pairs, while the filtered test set contains 1,199 papers and 8,531 QA pairs. This extensive collection of QA pairs aims to support the development and evaluation of open-ended QA systems in various scientific fields. This dataset has multiple applications: the training set can be used to fine-tune LLMs, infusing them with scientific knowledge; while the test set can be used to evaluate the overall performance of LLMs on open-ended, closed-book science QA tasks, or within specific domains. Given its substantial size, the test set can also serve as high-quality data for fine-tuning. Next, we provide several statistical metrics and visualizations to assess the quality of QA pairs from various perspectives.
\vspace{0.5ex}
\noindent
\\
\textbf{Source validation of numeric values}. We encourage the generator to reference data from the source paper when generating answers. However, we are concerned about inevitable hallucination \cite{xu2024hallucination}, wherein the model fabricates numbers as support. Given our requirement that both questions and answers stay within the scope of the source paper, we assume that a numerical value in a generated answer that does not occur in the original text has uncertain authenticity. Thus we define the following ratio, which ideally should have value 1:
\begin{equation}
\frac{{\text{{\# numeric values in answers that appear in paper}}}}{{\text{{total \# numeric values in answers}}}}
\label{eq:numbers}
\end{equation}

\begin{figure}[ht!]
\centering
\includegraphics[width=0.45\textwidth]{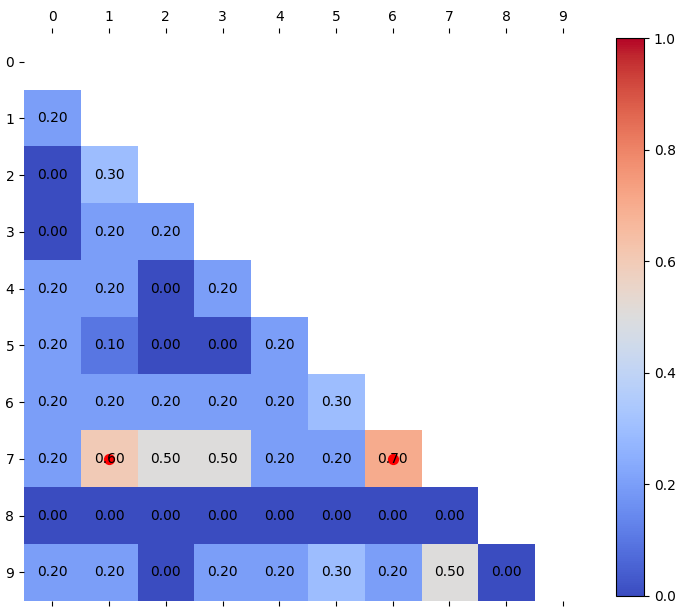}
\caption{Pairwise similarities between pairs of 10 questions generated from \citet{ivleva2009towards}. Lower similarity is indicated by bluer cells, while higher similarity is indicated by redder cells. Scores above 0.7 are marked with red dots.}
\label{fig:3}
\end{figure}

We find that 27.29\% of generated answers contain numerical information. Of the numeric values found in answers, 99.15\% occur in the source text, suggesting that fabricated numerical data are rare.
\vspace{0.5ex}
\noindent
\\
\textbf{Diversity of questions}: Since we aim for unique questions that are not mere paraphrases of each other, we calculated the text similarity between generated questions. Traditional metrics like METEOR \cite{banerjee2005meteor} fail to capture the true intent behind textually similar questions, as seen in the example:
\begin{itemize}
\setlength{\itemsep}{0em}
\item\textit{What are the advantages of using zeolites?}
\item\textit{What are the disadvantages of using zeolites?}
\end{itemize}
To address this, we use chain-of-thought (CoT) \cite{NEURIPS2022_9d560961} prompting with the Mixtral 8×7B model \cite{jiang2024mixtral} in a few-shot setting (see \autoref{sec:Appendix D}). It instructs the model to predict similarity scores by: 1) understanding the purpose; 2) identifying aspects; 3) comparing intent; and 4) assigning a score. We computed pairwise similarities for sets of 10 questions generated for each of 50 papers. Most questions exhibit dissimilarity (see \autoref{fig:3}). The distribution of the 2250 similarity scores, calculated from the 50×10×9/2 possible question pairs, reveals that 87.29\% of the scores fall below 0.3, while only 2.27\% exceed 0.7. The average similarity score of 0.186 demonstrates the effectiveness of the SciQAG method in generating a diverse set of questions with minimal overlap, ensuring that the generated questions are distinct from one another.

\vspace{0.1ex}
\noindent
\textbf{Coverage rate of answers}. To show our answers are from different parts of the paper, we measure the coverage rate of a paper's associated set of answers $A$. We first calculate the cosine similarity of each answer $a\in A$ to each sentence $s$ in the paper and then construct as $S_{15}$ the set of sentences in the paper that are in the \textit{topk} ($k$ = 15\%) similarity scores for any answer in $A$. We then divide the paper into 10 equally sized chunks and measure how many of these chunks contain any sentence from $S_{15}$. 
The coverage rate is defined as:
\begin{equation}
    P_{\text{cov\_rate}} = \frac{\sum_{i=1}^{10} \bigcup_{\substack{x \in topk(a) \\ a \in A}} \mathds{1}_{x}(C_i))}{100}
\end{equation}
where $C_i$ is the set of sentences in chunk $i$ of the paper ($i \in {1, 2, .., 10}$), $A$ is the set of generated answers, \textit{topk}(\textit{a}) is the set of 15\% of sentences in the paper that are the most similar to answer \textit{a}, and $\mathds{1}_x(A)$ is the indicator function indicating if $x \in A$. 

We calculate a similarity score between each answer and every sentence in the paper using contextual embeddings generated by the \texttt{e5-mistral-7b-instruct} model \cite{wang2024improving}. The average coverage rate for the 50 papers randomly selected is 78.6\%. Details of how sourcing works in practice are visualized in \autoref{sec:Appendix F}. 

\vspace{0.5ex}
\noindent
\section{Experiments}
% compare with zero-shot by prompting on 200 papers: LLaMA1, LLaMA2-chat, GPT-3.5, GPT-4, Claude3
% compare llama1 vs. finetune llama 1 on training set, further finetuning on single task & sciq.
\subsection{Experimental setup}
To investigate performance differences across various language models and the impact of fine-tuning, we designed a comprehensive experimental setup.
\vspace{0.5ex}
\noindent
\\
\textbf{Zero-shot setting}. We used part of the test set in SciQAG-24D to conduct a zero-shot comparison of five models. Two of them are open-sourced LLMs: \texttt{LLaMA}1 (7B) \cite{touvron2023llama1} and \texttt{LLaMA}2-chat (7B) \cite{touvron2023llama}. And the rest are commercial LLM systems (use API): GPT-3.5 (\texttt{gpt-3.5-turbo}) \cite{Ouyang2022}, GPT-4 (\texttt{gpt-4-1106-preview}) \cite{openai2023gpt4}, and Claude3 (\texttt{claude-3-opus-20240229}) \cite{anthropic_claude3}. Each model was prompted with 1000 questions in test, and their outputs were evaluated by CAR metrics (see \autoref{eval} for details) to gauge their zero-shot capabilities in answering scientific research questions.
\vspace{0.5ex}
\noindent
\\
\textbf{Fine-tuned setting}. Next, we examined the impact of fine-tuning. We fine-tuned \texttt{LLaMA}1 using the training set in SciQAG-24D. This fine-tuned model, \texttt{LLaMA}1-QA, was evaluated on an unseen test set in SciQAG-24D, and its performance was compared against the zero-shot performance of the original LLaMA1. Since the primary purpose of the training set is to fine-tune the model to infuse scientific knowledge, we further explored the performance of \texttt{LLaMA}1-QA after fine-tuning on other scientific tasks. The first task was SciQ \cite{welbl-etal-2017-crowdsourcing}, a scientific MCQ benchmark and we used 10,000 examples as training data and 1,000 examples as test data. We also conducted evaluations on several other scientific tasks (see \autoref{table:dataset} for details), including different types like classification, regression, and transformation.

\subsection{Implementation details}
For zero-shot experiments, we set repetition penalty of \texttt{LLaMA}1 to 1.2 and max token length of \texttt{LLaMA}2-chat to 1024. Apart from these specific settings, the rest are configured with repetition set to 1 and the max token length set to 512. To fine-tune the \texttt{LLaMA}1 model on QA pairs, we follow the method of Alpaca \cite{alpaca}, using a batch size of 1 and a maximum sequence length of 512 tokens. The training process is conducted over 2 epochs using the DeepSpeed Stage 3 optimization framework across 8 AMD GPUs. Gradients are accumulated over 2 steps, and the learning rate is set to 2e-5 with no weight decay. The number of warm-up steps is set to 20. For all inference, we set the temperature to 0.8 and the top\_p to 0.75. 

\subsection{Evaluation metrics}
\label{eval}
For the evaluation of the quality of generated answers in the zero-shot experiment, we adopt a modified version of the RACAR framework. We exclude the criteria of "Relevance" and "Agnosticism", which are specifically designed to assess the alignment of question-answer (QA) pairs with a given paper, and name this modified metric as \textbf{CAR} (see \autoref{sec:Appendix I}). For evaluating performance on SciQ dataset, we utilize the accuracy metric to conduct our evaluations. For evaluating performance on other scientific tasks, we use F1-score for classification tasks, Mean Absolute Error (MAE) for regression tasks, and KL divergence \cite{brown2019guacamol} for transformation tasks.

\begin{table*}[ht!]
\centering
\begin{adjustbox}{width=0.8\textwidth}
\begin{tabular}{l|ccc}
\hline
\textbf{Model} & \textbf{C}ompleteness & \textbf{A}ccuracy & \textbf{R}easonableness\\
\hline
\texttt{LLaMA}1 & 2.7813 & 2.9625 & 3.2917 \\
\texttt{LLaMA}2-chat & 4.7125 & 4.4708 & 4.7625\\
\texttt{GPT-3.5} & 4.8708 & 4.9333 & 4.9895\\
\texttt{GPT-4} & \textbf{4.9} & 4.9438  & \textbf{4.9896} \\
\texttt{Claude3}  & 4.7938 & \textbf{4.95} & 4.9875\\
\hline
\texttt{LLAMA}1-QA & 3.1479 (+13.18\%) & 3.8875 (+31.22\%) & 4.3896 (+33.35\%)\\
\hline
\end{tabular}
\end{adjustbox}
\caption{Zero-shot performance on SciQAG-24D test. \textbf{Bold} denotes best model scores. As a comparison with \texttt{LLaMA}1, we also show results of \texttt{LLaMA}1\_QA in the last line.}
\label{table:3}
\end{table*}

\begin{table}[bt!]
\begin{tabular}{llcc}
\hline
Type & Dataset & \texttt{LLaMA}1 & \texttt{LLAMA}1-QA \\
\hline
M & \texttt{SciQ} & 0.895 & 0.904 \\
\hline
\multirow{4}{1pt}{C} & \texttt{Glass} & 0.813 & 0.872 \\
& \texttt{Pei} & 0.9531 & 0.960 \\
& \texttt{Stability} & 0.809 & 0.931 \\
& \texttt{UV} & 0.596 & 0.729 \\
\hline
\multirow{4}{1pt}{R} & \texttt{Bandgap} & 0.435 & 0.384 \\
& \texttt{Steels} & 463.959 & 185.316 \\
& \texttt{ChEMBL} & 1.195 & 1.180 \\
& \texttt{Diversity} & 0.307 & 0.294 \\
& \texttt{Cp} & 4.790 & 4.645 \\
\hline
\multirow{2}{*}{T} & \texttt{Stability} & 0.647 & 0.824 \\
& \texttt{ESOL} & 0.788 & 0.829 \\
\hline
\end{tabular}
\caption{Fine-tuning performance of \texttt{LLaMA}1 and \texttt{LLAMA}1-QA on SciQ and scientific tasks (C for classification; R for regression; T for transformation). The details of datasets are available in \autoref{table:dataset}.}
\label{table:4}
\end{table}

\section{Results and Analysis}
\label{sec:results}
\subsection{Zero-shot Evaluation} 
The results of the zero-shot performance on the SciQAG-24D test are summarized in Table \ref{table:3}. Among the models, \texttt{GPT-4} achieved the highest scores in Completeness (4.9) and Reasonableness (4.9896), while \texttt{Claude3} obtained the highest Accuracy score (4.95). \texttt{GPT-3.5} also performed exceptionally well, with scores closely trailing those of \texttt{GPT-4} and Claude3 across all metrics. Notably, \texttt{LLaMA}1 scores the lowest in all three dimensions. In comparison, the \texttt{LLaMA}2-chat model, although not as high-scoring as the GPT models, demonstrates substantial improvements over the original \texttt{LLaMA}1 in all metrics. These results highlight the superior zero-shot performance of the commercial LLM models in answering science questions. We can also find that the latest open-source models, such as \texttt{LLaMA}2-chat, have made significant advancements and possess science question-answering capabilities comparable to commercial LLMs.

\subsection{Impact of Fine-tuning} 
We selected \texttt{LLaMA1}, which exhibited the poorest zero-shot performance, for fine-tuning to obtain \texttt{LLaMA}1-QA. Through three experiments, we demonstrated that SciQAG-24D serves as effective fine-tuning data capable of improving downstream scientific task performance. 
\vspace{0.5ex}
\noindent
\\
\textbf{SciQAG-24D test}. As shown in the last line of \autoref{table:3}, \texttt{LLaMA}1-QA, demonstrated significant performance improvement compared to the original \texttt{LLaMA}1 (over 13\% in Completeness, and over 30\% in Accuracy and Reasonableness). This indicates that \texttt{LLaMA}1 has learned the logic of answering science questions from the training data in SciQAG-24D and  internalized some scientific knowledge. 
\vspace{0.5ex}
\noindent
\\
\textbf{SciQ test}. In the first line of \autoref{table:4}, \texttt{LLaMA}1-QA exhibits a slight improvement over \texttt{LLaMA}1 (+1\%). However, based on our observations and statistics, fine-tuning on SciQAG-24D has also enhanced the model's instruction-following ability: the SciQ test requires selecting an option as an answer, but occasionally, LLMs may generate format errors that prevent automatic parsing of the predicted option from the output. The probability of unparseable outputs has decreased from 4.1\% with \texttt{LLaMA}1 to 1.7\% with \texttt{LLaMA}1-QA.
\vspace{0.5ex}
\noindent
\\
\textbf{Scientific tasks}. In \autoref{table:4}, \texttt{LLaMA}1-QA exhibits notable improvements over \texttt{LLaMA}1 model in  scientific tasks. The most significant improvement is observed in the \texttt{Steels} regression task, where the MAE decreased from 463.96 to 185.32. These findings indicate that incorporating science QA into the training process can enhance the model's ability to learn scientific knowledge, thereby improving its performance on downstream prediction tasks. Surprisingly, compared to machine learning models specifically designed, LLM can achieve comparable results or even surpass them in certain tasks. For example, in the Bandgap task, although \texttt{LLaMA}1-QA does not perform as well as models like MODNet (0.3327), it has already surpassed AMMExpress v2020 (0.4161) \cite{Dunn2020}. In the Diversity task, \texttt{LLaMA}1-QA outperforms the deep learning baseline (0.3198) \cite{jablonka2024leveraging}. These findings demonstrates that LLMs hold significant potential for specific scientific tasks.

\section{Conclusion}
In this work, we present SciQAG, a novel framework for generating QA pairs from scientific literature, and propose RACAR metrics for evaluating and filtering these QA pairs. SciQAG can cost-effectively produce knowledge-based QA data for low-resource scientific domains, addressing a key challenge in developing science QA systems. Using SciQAG, we create an science QA dataset with 188,042 QA pairs. Its training set can fine-tune LLMs with scientific knowledge, and its test set can evaluate LLM performance on open-ended, closed-book science QA tasks. To assess the impact of the dataset, we compare the zero-shot performance of several LLMs on the test set and fine-tune \texttt{LLaMA}1 on the training set, resulting in the \texttt{LLaMA}1-QA model. This fine-tuning significantly improves \texttt{LLaMA}1-QA's performance on the test set (unseen QA), the SciQ test (MCQ), and various scientific tasks (real-world experimental problems). Additionally, this study demonstrates the potential of LLMs in scientific tasks, with \texttt{LLaMA}1-QA achieving results comparable to or surpassing machine learning baselines. These findings highlight the utility of the SciQAG-24D dataset in enhancing LLMs' understanding and application of scientific knowledge. We aim to promote future research in this field, advancing science QA systems and providing valuable resources for practitioners.

\section*{Limitations}
Firstly, although the questions in the dataset do not explicitly depend on the original text, there may still be implicit dependencies, such as author-specific terminology or names. This can lead to gap between what the LLM learns during training and real knowledge embedded in the QA pairs. Secondly, hallucinations, while relatively infrequent, are still inevitable due to the inherent nature of LLMs. Even a perfect RACAR score does not guarantee that the generated answer is entirely accurate or factual. Thirdly, the RACAR evaluation framework appears to perform better when using commercial models compared to open-source models, and when assessing each dimension individually rather than all five dimensions simultaneously. However, this approach is more resource-intensive. Future work should explore the balance between computational efficiency and performance to develop more sustainable evaluation strategies. Lastly, the current version of the SciQAG-24D dataset covers a limited range of scientific domains. To enhance the dataset's representativeness and utility, it is crucial for the research community to collaborate and contribute to its expansion, incorporating a wider variety of scientific disciplines and question types.

\section*{Ethics Statement}

Our research does not raise any major ethical concerns.

%\section*{Acknowledgements}
%This work has been funded by Microsoft.

\bibliography{acl_latex}

\begin{thebibliography}{83}
\expandafter\ifx\csname natexlab\endcsname\relax\def\natexlab#1{#1}\fi

\bibitem[{Anthropic(2024)}]{anthropic_claude3}
Anthropic. 2024.
\newblock The claude 3 model family: Opus, sonnet, haiku.
\newblock Retrieved from \url{https://www-cdn.anthropic.com/de8ba9b01c9ab7cbabf5c33b80b7bbc618857627/Model_Card_Claude_3.pdf}.

\bibitem[{Auer et~al.(2023)Auer, Barone, Bartz, Cortes, Jaradeh, Karras, Koubarakis, Mouromtsev, Pliukhin, Radyush et~al.}]{auer2023sciqa}
S{\"o}ren Auer, Dante~AC Barone, Cassiano Bartz, Eduardo~G Cortes, Mohamad~Yaser Jaradeh, Oliver Karras, Manolis Koubarakis, Dmitry Mouromtsev, Dmitrii Pliukhin, Daniil Radyush, et~al. 2023.
\newblock The sciqa scientific question answering benchmark for scholarly knowledge.
\newblock \emph{Scientific Reports}, 13(1):7240.

\bibitem[{Banerjee and Lavie(2005{\natexlab{a}})}]{banerjee-lavie-2005-meteor}
Satanjeev Banerjee and Alon Lavie. 2005{\natexlab{a}}.
\newblock \href {https://aclanthology.org/W05-0909} {{METEOR}: An automatic metric for {MT} evaluation with improved correlation with human judgments}.
\newblock In \emph{Proceedings of the {ACL} Workshop on Intrinsic and Extrinsic Evaluation Measures for Machine Translation and/or Summarization}, pages 65--72, Ann Arbor, Michigan. Association for Computational Linguistics.

\bibitem[{Banerjee and Lavie(2005{\natexlab{b}})}]{banerjee2005meteor}
Satanjeev Banerjee and Alon Lavie. 2005{\natexlab{b}}.
\newblock Meteor: An automatic metric for mt evaluation with improved correlation with human judgments.
\newblock In \emph{Proceedings of the acl workshop on intrinsic and extrinsic evaluation measures for machine translation and/or summarization}, pages 65--72.

\bibitem[{Batra et~al.(2020)Batra, Chen, Evans, Walton, and Ramprasad}]{Batra2020}
Rohit Batra, Carmen Chen, Tania~G Evans, Krista~S Walton, and Rampi Ramprasad. 2020.
\newblock {Prediction of water stability of metal–organic frameworks using machine learning}.
\newblock \emph{Nature Machine Intelligence}, 2(11):704--710.

\bibitem[{Beard et~al.(2019)Beard, Sivaraman, Vázquez-Mayagoitia, Vishwanath, and Cole}]{beard_comparative_2019}
Edward~J. Beard, Ganesh Sivaraman, Álvaro Vázquez-Mayagoitia, Venkatram Vishwanath, and Jacqueline~M. Cole. 2019.
\newblock \href {https://doi.org/10.1038/s41597-019-0306-0} {Comparative dataset of experimental and computational attributes of {UV}/vis absorption spectra}.
\newblock \emph{Scientific Data}, 6(1):307.
\newblock Publisher: Nature Publishing Group.

\bibitem[{Brown et~al.(2019)Brown, Fiscato, Segler, and Vaucher}]{brown2019guacamol}
Nathan Brown, Marco Fiscato, Marwin~HS Segler, and Alain~C Vaucher. 2019.
\newblock Guacamol: benchmarking models for de novo molecular design.
\newblock \emph{Journal of chemical information and modeling}, 59(3):1096--1108.

\bibitem[{Chen et~al.(2019)Chen, Stanovsky, Singh, and Gardner}]{chen-etal-2019-evaluating}
Anthony Chen, Gabriel Stanovsky, Sameer Singh, and Matt Gardner. 2019.
\newblock \href {https://doi.org/10.18653/v1/D19-5817} {Evaluating question answering evaluation}.
\newblock In \emph{Proceedings of the 2nd Workshop on Machine Reading for Question Answering}, pages 119--124, Hong Kong, China. Association for Computational Linguistics.

\bibitem[{Chen et~al.(2017)Chen, Fisch, Weston, and Bordes}]{chen2017reading}
Danqi Chen, Adam Fisch, Jason Weston, and Antoine Bordes. 2017.
\newblock Reading wikipedia to answer open-domain questions.
\newblock \emph{arXiv preprint arXiv:1704.00051}.

\bibitem[{Chen et~al.(2023)Chen, Qian, Tang, Lai, Liu, Han, and Jia}]{chen2023longlora}
Yukang Chen, Shengju Qian, Haotian Tang, Xin Lai, Zhijian Liu, Song Han, and Jiaya Jia. 2023.
\newblock \href {http://arxiv.org/abs/2309.12307} {Long{LoRA}: Efficient fine-tuning of long-context large language models}.

\bibitem[{Chowdhery et~al.(2022)Chowdhery, Narang, Devlin, Bosma, Mishra, Roberts, Barham, Chung, Sutton, Gehrmann, Schuh, Shi, Tsvyashchenko, Maynez, Rao, Barnes, Tay, Shazeer, Prabhakaran, Reif, Du, Hutchinson, Pope, Bradbury, Austin, Isard, Gur-Ari, Yin, Duke, Levskaya, Ghemawat, Dev, Michalewski, Garcia, Misra, Robinson, Fedus, Zhou, Ippolito, Luan, Lim, Zoph, Spiridonov, Sepassi, Dohan, Agrawal, Omernick, Dai, Pillai, Pellat, Lewkowycz, Moreira, Child, Polozov, Lee, Zhou, Wang, Saeta, Diaz, Firat, Catasta, Wei, Meier-Hellstern, Eck, Dean, Petrov, and Fiedel}]{chowdhery2022palm}
Aakanksha Chowdhery, Sharan Narang, Jacob Devlin, Maarten Bosma, Gaurav Mishra, Adam Roberts, Paul Barham, Hyung~Won Chung, Charles Sutton, Sebastian Gehrmann, Parker Schuh, Kensen Shi, Sasha Tsvyashchenko, Joshua Maynez, Abhishek Rao, Parker Barnes, Yi~Tay, Noam Shazeer, Vinodkumar Prabhakaran, Emily Reif, Nan Du, Ben Hutchinson, Reiner Pope, James Bradbury, Jacob Austin, Michael Isard, Guy Gur-Ari, Pengcheng Yin, Toju Duke, Anselm Levskaya, Sanjay Ghemawat, Sunipa Dev, Henryk Michalewski, Xavier Garcia, Vedant Misra, Kevin Robinson, Liam Fedus, Denny Zhou, Daphne Ippolito, David Luan, Hyeontaek Lim, Barret Zoph, Alexander Spiridonov, Ryan Sepassi, David Dohan, Shivani Agrawal, Mark Omernick, Andrew~M. Dai, Thanumalayan~Sankaranarayana Pillai, Marie Pellat, Aitor Lewkowycz, Erica Moreira, Rewon Child, Oleksandr Polozov, Katherine Lee, Zongwei Zhou, Xuezhi Wang, Brennan Saeta, Mark Diaz, Orhan Firat, Michele Catasta, Jason Wei, Kathy Meier-Hellstern, Douglas Eck, Jeff Dean, Slav Petrov, and Noah Fiedel. 2022.
\newblock \href {http://arxiv.org/abs/2204.02311} {Pa{LM}: Scaling language modeling with pathways}.

\bibitem[{Clark et~al.(2019)Clark, Celikyilmaz, and Smith}]{clark-etal-2019-sentence}
Elizabeth Clark, Asli Celikyilmaz, and Noah~A. Smith. 2019.
\newblock \href {https://doi.org/10.18653/v1/P19-1264} {Sentence mover{'}s similarity: Automatic evaluation for multi-sentence texts}.
\newblock In \emph{Proceedings of the 57th Annual Meeting of the Association for Computational Linguistics}, pages 2748--2760, Florence, Italy. Association for Computational Linguistics.

\bibitem[{Clark et~al.(2018)Clark, Cowhey, Etzioni, Khot, Sabharwal, Schoenick, and Tafjord}]{clark2018think}
Peter Clark, Isaac Cowhey, Oren Etzioni, Tushar Khot, Ashish Sabharwal, Carissa Schoenick, and Oyvind Tafjord. 2018.
\newblock Think you have solved question answering? try arc, the ai2 reasoning challenge.
\newblock \emph{arXiv preprint arXiv:1803.05457}.

\bibitem[{Dao(2023)}]{dao2023flashattention2}
Tri Dao. 2023.
\newblock \href {http://arxiv.org/abs/2307.08691} {Flash{A}ttention-2: Faster attention with better parallelism and work partitioning}.

\bibitem[{Das et~al.(2021)Das, Majumder, Phadikar, and Sekh}]{das2021automatic}
Bidyut Das, Mukta Majumder, Santanu Phadikar, and Arif~Ahmed Sekh. 2021.
\newblock \href {https://doi.org/10.1186/s41039-021-00151-1} {Automatic question generation and answer assessment: A survey}.
\newblock \emph{Research and Practice in Technology Enhanced Learning}, 16(1):1--15.

\bibitem[{Dasigi et~al.(2021)Dasigi, Lo, Beltagy, Cohan, Smith, and Gardner}]{dasigi-etal-2021-dataset}
Pradeep Dasigi, Kyle Lo, Iz~Beltagy, Arman Cohan, Noah~A. Smith, and Matt Gardner. 2021.
\newblock \href {https://doi.org/10.18653/v1/2021.naacl-main.365} {A dataset of information-seeking questions and answers anchored in research papers}.
\newblock In \emph{Proceedings of the 2021 Conference of the North American Chapter of the Association for Computational Linguistics: Human Language Technologies}, pages 4599--4610, Online. Association for Computational Linguistics.

\bibitem[{Delaney(2004)}]{Delaney2004}
John~S Delaney. 2004.
\newblock {ESOL: estimating aqueous solubility directly from molecular structure}.
\newblock \emph{Journal of chemical information and computer sciences}, 44(3):1000--1005.

\bibitem[{Dunn et~al.(2020)Dunn, Wang, Ganose, Dopp, and Jain}]{Dunn2020}
Alexander Dunn, Qi~Wang, Alex Ganose, Daniel Dopp, and Anubhav Jain. 2020.
\newblock {Benchmarking materials property prediction methods: the Matbench test set and Automatminer reference algorithm}.
\newblock \emph{npj Computational Materials}, 6(1):138.

\bibitem[{Fang et~al.(2024)Fang, Liang, Zhang, Liu, Huang, Chen, Fan, and Chen}]{fang2024molinstructions}
Yin Fang, Xiaozhuan Liang, Ningyu Zhang, Kangwei Liu, Rui Huang, Zhuo Chen, Xiaohui Fan, and Huajun Chen. 2024.
\newblock \href {https://openreview.net/forum?id=Tlsdsb6l9n} {Mol-{I}nstructions -- {A} large-scale biomolecular instruction dataset for large language models}.
\newblock In \emph{12th International Conference on Learning Representations}.

\bibitem[{Gaulton et~al.(2012)Gaulton, Bellis, Bento, Chambers, Davies, Hersey, Light, McGlinchey, Michalovich, and Al-Lazikani}]{Gaulton2012}
Anna Gaulton, Louisa~J Bellis, A~Patricia Bento, Jon Chambers, Mark Davies, Anne Hersey, Yvonne Light, Shaun McGlinchey, David Michalovich, and Bissan Al-Lazikani. 2012.
\newblock {ChEMBL: a large-scale bioactivity database for drug discovery}.
\newblock \emph{Nucleic acids research}, 40(D1):D1100--D1107.

\bibitem[{Heilman and Smith(2010)}]{heilman2010good}
Michael Heilman and Noah~A Smith. 2010.
\newblock Good question! statistical ranking for question generation.
\newblock In \emph{Human language technologies: The 2010 annual conference of the North American Chapter of the Association for Computational Linguistics}, pages 609--617.

\bibitem[{Hendrycks et~al.(2020)Hendrycks, Burns, Basart, Zou, Mazeika, Song, and Steinhardt}]{hendrycks2020measuring}
Dan Hendrycks, Collin Burns, Steven Basart, Andy Zou, Mantas Mazeika, Dawn Song, and Jacob Steinhardt. 2020.
\newblock Measuring massive multitask language understanding.
\newblock \emph{arXiv preprint arXiv:2009.03300}.

\bibitem[{Hermann et~al.(2015)Hermann, Kocisky, Grefenstette, Espeholt, Kay, Suleyman, and Blunsom}]{hermann2015teaching}
Karl~Moritz Hermann, Tomas Kocisky, Edward Grefenstette, Lasse Espeholt, Will Kay, Mustafa Suleyman, and Phil Blunsom. 2015.
\newblock Teaching machines to read and comprehend.
\newblock \emph{Advances in neural information processing systems}, 28.

\bibitem[{Ivleva et~al.(2009)Ivleva, Wagner, Horn, Niessner, and Haisch}]{ivleva2009towards}
Natalia~P Ivleva, Michael Wagner, Harald Horn, Reinhard Niessner, and Christoph Haisch. 2009.
\newblock Towards a nondestructive chemical characterization of biofilm matrix by raman microscopy.
\newblock \emph{Analytical and bioanalytical chemistry}, 393:197--206.

\bibitem[{Jablonka et~al.(2024)Jablonka, Schwaller, Ortega-Guerrero, and Smit}]{jablonka2024leveraging}
Kevin~Maik Jablonka, Philippe Schwaller, Andres Ortega-Guerrero, and Berend Smit. 2024.
\newblock Leveraging large language models for predictive chemistry.
\newblock \emph{Nature Machine Intelligence}, pages 1--9.

\bibitem[{Jiang et~al.(2024)Jiang, Sablayrolles, Roux, Mensch, Savary, Bamford, Chaplot, de~las Casas, Hanna, Bressand, Lengyel, Bour, Lample, Lavaud, Saulnier, Lachaux, Stock, Subramanian, Yang, Antoniak, Scao, Gervet, Lavril, Wang, Lacroix, and Sayed}]{jiang2024mixtral}
Albert~Q. Jiang, Alexandre Sablayrolles, Antoine Roux, Arthur Mensch, Blanche Savary, Chris Bamford, Devendra~Singh Chaplot, Diego de~las Casas, Emma~Bou Hanna, Florian Bressand, Gianna Lengyel, Guillaume Bour, Guillaume Lample, Lélio~Renard Lavaud, Lucile Saulnier, Marie-Anne Lachaux, Pierre Stock, Sandeep Subramanian, Sophia Yang, Szymon Antoniak, Teven~Le Scao, Théophile Gervet, Thibaut Lavril, Thomas Wang, Timothée Lacroix, and William~El Sayed. 2024.
\newblock \href {http://arxiv.org/abs/2401.04088} {Mixtral of experts}.

\bibitem[{Jin et~al.(2021)Jin, Pan, Oufattole, Weng, Fang, and Szolovits}]{jin2020disease}
Di~Jin, Eileen Pan, Nassim Oufattole, Wei-Hung Weng, Hanyi Fang, and Peter Szolovits. 2021.
\newblock \href {https://doi.org/10.3390/app11146421} {What disease does this patient have? {A} large-scale open domain question answering dataset from medical exams}.
\newblock \emph{Applied Sciences}, 11(14).

\bibitem[{Jin et~al.(2019)Jin, Dhingra, Liu, Cohen, and Lu}]{jin-etal-2019-pubmedqa}
Qiao Jin, Bhuwan Dhingra, Zhengping Liu, William Cohen, and Xinghua Lu. 2019.
\newblock \href {https://doi.org/10.18653/v1/D19-1259} {{P}ub{M}ed{QA}: A dataset for biomedical research question answering}.
\newblock In \emph{Proceedings of the 2019 Conference on Empirical Methods in Natural Language Processing and the 9th International Joint Conference on Natural Language Processing (EMNLP-IJCNLP)}, pages 2567--2577, Hong Kong, China. Association for Computational Linguistics.

\bibitem[{Kamalloo et~al.(2023)Kamalloo, Dziri, Clarke, and Rafiei}]{kamalloo-etal-2023-evaluating}
Ehsan Kamalloo, Nouha Dziri, Charles Clarke, and Davood Rafiei. 2023.
\newblock \href {https://doi.org/10.18653/v1/2023.acl-long.307} {Evaluating open-domain question answering in the era of large language models}.
\newblock In \emph{Proceedings of the 61st Annual Meeting of the Association for Computational Linguistics (Volume 1: Long Papers)}, pages 5591--5606, Toronto, Canada. Association for Computational Linguistics.

\bibitem[{Kawazoe et~al.(1997)Kawazoe, Yu, Tsai, and Masumoto}]{Kawazoe1997}
Yoshiyuki Kawazoe, J-Z Yu, A-P Tsai, and T~Masumoto. 1997.
\newblock {Nonequilibrium phase diagrams of ternary amorphous alloys}.

\bibitem[{Krska and Molinelli(2009)}]{krska2009rapid}
Rudolf Krska and Alexandra Molinelli. 2009.
\newblock Rapid test strips for analysis of mycotoxins in food and feed.
\newblock \emph{Analytical and bioanalytical chemistry}, 393:67--71.

\bibitem[{Lee et~al.(2020)Lee, Yoon, Kim, Kim, Kim, So, and Kang}]{lee2020biobert}
Jinhyuk Lee, Wonjin Yoon, Sungdong Kim, Donghyeon Kim, Sunkyu Kim, Chan~Ho So, and Jaewoo Kang. 2020.
\newblock Biobert: a pre-trained biomedical language representation model for biomedical text mining.
\newblock \emph{Bioinformatics}, 36(4):1234--1240.

\bibitem[{Lee et~al.(2023{\natexlab{a}})Lee, Kim, and Kang}]{lee2023liquid}
Seongyun Lee, Hyunjae Kim, and Jaewoo Kang. 2023{\natexlab{a}}.
\newblock \href {https://doi.org/10.48550/arXiv.2302.01691} {{LIQUID}: A framework for list question answering dataset generation}.
\newblock \emph{arXiv preprint arXiv:2302.01691}.

\bibitem[{Lee et~al.(2023{\natexlab{b}})Lee, Lee, Park, Hwang, Kim, Lee, and Lee}]{lee2023qasa}
Yoonjoo Lee, Kyungjae Lee, Sunghyun Park, Dasol Hwang, Jaehyeon Kim, Hong-in Lee, and Moontae Lee. 2023{\natexlab{b}}.
\newblock {QASA}: Advanced question answering on scientific articles.
\newblock In \emph{Proceedings of the 40th International Conference on Machine Learning}.

\bibitem[{Lin(2004)}]{lin-2004-rouge}
Chin-Yew Lin. 2004.
\newblock \href {https://aclanthology.org/W04-1013} {{ROUGE}: A package for automatic evaluation of summaries}.
\newblock In \emph{Text Summarization Branches Out}, pages 74--81, Barcelona, Spain. Association for Computational Linguistics.

\bibitem[{Longpre et~al.(2021)Longpre, Lu, and Daiber}]{longpre2021mkqa}
Shayne Longpre, Yi~Lu, and Joachim Daiber. 2021.
\newblock Mkqa: A linguistically diverse benchmark for multilingual open domain question answering.
\newblock \emph{Transactions of the Association for Computational Linguistics}, 9:1389--1406.

\bibitem[{Lu et~al.(2022)Lu, Mishra, Xia, Qiu, Chang, Zhu, Tafjord, Clark, and Kalyan}]{lu2022learn}
Pan Lu, Swaroop Mishra, Tony Xia, Liang Qiu, Kai-Wei Chang, Song-Chun Zhu, Oyvind Tafjord, Peter Clark, and Ashwin Kalyan. 2022.
\newblock \href {https://openreview.net/forum?id=HjwK-Tc_Bc} {Learn to explain: Multimodal reasoning via thought chains for science question answering}.
\newblock In \emph{Advances in Neural Information Processing Systems}.

\bibitem[{Masry et~al.(2022)Masry, Long, Tan, Joty, and Hoque}]{masry2022chartqa}
Ahmed Masry, Do~Xuan Long, Jia~Qing Tan, Shafiq Joty, and Enamul Hoque. 2022.
\newblock Chartqa: A benchmark for question answering about charts with visual and logical reasoning.
\newblock \emph{arXiv preprint arXiv:2203.10244}.

\bibitem[{Mihaylov et~al.(2018)Mihaylov, Clark, Khot, and Sabharwal}]{mihaylov-etal-2018-suit}
Todor Mihaylov, Peter Clark, Tushar Khot, and Ashish Sabharwal. 2018.
\newblock \href {https://doi.org/10.18653/v1/D18-1260} {Can a suit of armor conduct electricity? {A} new dataset for open book question answering}.
\newblock In \emph{Proceedings of the 2018 Conference on Empirical Methods in Natural Language Processing}, pages 2381--2391, Brussels, Belgium. Association for Computational Linguistics.

\bibitem[{Min et~al.(2023)Min, Krishna, Lyu, Lewis, Yih, Koh, Iyyer, Zettlemoyer, and Hajishirzi}]{min-etal-2023-factscore}
Sewon Min, Kalpesh Krishna, Xinxi Lyu, Mike Lewis, Wen-tau Yih, Pang Koh, Mohit Iyyer, Luke Zettlemoyer, and Hannaneh Hajishirzi. 2023.
\newblock \href {https://doi.org/10.18653/v1/2023.emnlp-main.741} {{FA}ct{S}core: Fine-grained atomic evaluation of factual precision in long form text generation}.
\newblock In \emph{Proceedings of the 2023 Conference on Empirical Methods in Natural Language Processing}, pages 12076--12100, Singapore. Association for Computational Linguistics.

\bibitem[{Mitkov et~al.(2003)}]{mitkov2003computer}
Ruslan Mitkov et~al. 2003.
\newblock Computer-aided generation of multiple-choice tests.
\newblock In \emph{Proceedings of the HLT-NAACL 03 workshop on Building educational applications using natural language processing}, pages 17--22.

\bibitem[{Moosavi et~al.(2020)Moosavi, Nandy, Jablonka, Ongari, Janet, Boyd, Lee, Smit, and Kulik}]{Moosavi2020}
Seyed~Mohamad Moosavi, Aditya Nandy, Kevin~Maik Jablonka, Daniele Ongari, Jon~Paul Janet, Peter~G Boyd, Yongjin Lee, Berend Smit, and Heather~J Kulik. 2020.
\newblock {Understanding the diversity of the metal-organic framework ecosystem}.
\newblock \emph{Nature communications}, 11(1):1--10.

\bibitem[{Mulla and Gharpure(2023)}]{mulla_automatic_2023}
Nikahat Mulla and Prachi Gharpure. 2023.
\newblock \href {https://doi.org/10.1007/s13748-023-00295-9} {Automatic question generation: A review of methodologies, datasets, evaluation metrics, and applications}.
\newblock \emph{Progress in Artificial Intelligence}, 12(1):1--32.

\bibitem[{OpenAI et~al.(2023)OpenAI, :, Achiam, Adler, Agarwal, Ahmad, Akkaya, Aleman, Almeida, Altenschmidt, Altman, Anadkat, Avila, Babuschkin, Balaji, Balcom, Baltescu, Bao, Bavarian, Belgum, Bello, Berdine, Bernadett-Shapiro, Berner, Bogdonoff, Boiko, Boyd, Brakman, Brockman, Brooks, Brundage, Button, Cai, Campbell, Cann, Carey, Carlson, Carmichael, Chan, Chang, Chantzis, Chen, Chen, Chen, Chen, Chen, Chess, Cho, Chu, Chung, Cummings, Currier, Dai, Decareaux, Degry, Deutsch, Deville, Dhar, Dohan, Dowling, Dunning, Ecoffet, Eleti, Eloundou, Farhi, Fedus, Felix, Fishman, Forte, Fulford, Gao, Georges, Gibson, Goel, Gogineni, Goh, Gontijo-Lopes, Gordon, Grafstein, Gray, Greene, Gross, Gu, Guo, Hallacy, Han, Harris, He, Heaton, Heidecke, Hesse, Hickey, Hickey, Hoeschele, Houghton, Hsu, Hu, Hu, Huizinga, Jain, Jain, Jang, Jiang, Jiang, Jin, Jin, Jomoto, Jonn, Jun, Kaftan, Łukasz Kaiser, Kamali, Kanitscheider, Keskar, Khan, Kilpatrick, Kim, Kim, Kim, Kirchner, Kiros, Knight, Kokotajlo, Łukasz Kondraciuk,
  Kondrich, Konstantinidis, Kosic, Krueger, Kuo, Lampe, Lan, Lee, Leike, Leung, Levy, Li, Lim, Lin, Lin, Litwin, Lopez, Lowe, Lue, Makanju, Malfacini, Manning, Markov, Markovski, Martin, Mayer, Mayne, McGrew, McKinney, McLeavey, McMillan, McNeil, Medina, Mehta, Menick, Metz, Mishchenko, Mishkin, Monaco, Morikawa, Mossing, Mu, Murati, Murk, Mély, Nair, Nakano, Nayak, Neelakantan, Ngo, Noh, Ouyang, O'Keefe, Pachocki, Paino, Palermo, Pantuliano, Parascandolo, Parish, Parparita, Passos, Pavlov, Peng, Perelman, de~Avila Belbute~Peres, Petrov, de~Oliveira~Pinto, Michael, Pokorny, Pokrass, Pong, Powell, Power, Power, Proehl, Puri, Radford, Rae, Ramesh, Raymond, Real, Rimbach, Ross, Rotsted, Roussez, Ryder, Saltarelli, Sanders, Santurkar, Sastry, Schmidt, Schnurr, Schulman, Selsam, Sheppard, Sherbakov, Shieh, Shoker, Shyam, Sidor, Sigler, Simens, Sitkin, Slama, Sohl, Sokolowsky, Song, Staudacher, Such, Summers, Sutskever, Tang, Tezak, Thompson, Tillet, Tootoonchian, Tseng, Tuggle, Turley, Tworek, Uribe, Vallone,
  Vijayvergiya, Voss, Wainwright, Wang, Wang, Wang, Ward, Wei, Weinmann, Welihinda, Welinder, Weng, Weng, Wiethoff, Willner, Winter, Wolrich, Wong, Workman, Wu, Wu, Wu, Xiao, Xu, Yoo, Yu, Yuan, Zaremba, Zellers, Zhang, Zhang, Zhao, Zheng, Zhuang, Zhuk, and Zoph}]{openai2023gpt4}
OpenAI, :, Josh Achiam, Steven Adler, Sandhini Agarwal, Lama Ahmad, Ilge Akkaya, Florencia~Leoni Aleman, Diogo Almeida, Janko Altenschmidt, Sam Altman, Shyamal Anadkat, Red Avila, Igor Babuschkin, Suchir Balaji, Valerie Balcom, Paul Baltescu, Haiming Bao, Mo~Bavarian, Jeff Belgum, Irwan Bello, Jake Berdine, Gabriel Bernadett-Shapiro, Christopher Berner, Lenny Bogdonoff, Oleg Boiko, Madelaine Boyd, Anna-Luisa Brakman, Greg Brockman, Tim Brooks, Miles Brundage, Kevin Button, Trevor Cai, Rosie Campbell, Andrew Cann, Brittany Carey, Chelsea Carlson, Rory Carmichael, Brooke Chan, Che Chang, Fotis Chantzis, Derek Chen, Sully Chen, Ruby Chen, Jason Chen, Mark Chen, Ben Chess, Chester Cho, Casey Chu, Hyung~Won Chung, Dave Cummings, Jeremiah Currier, Yunxing Dai, Cory Decareaux, Thomas Degry, Noah Deutsch, Damien Deville, Arka Dhar, David Dohan, Steve Dowling, Sheila Dunning, Adrien Ecoffet, Atty Eleti, Tyna Eloundou, David Farhi, Liam Fedus, Niko Felix, Simón~Posada Fishman, Juston Forte, Isabella Fulford, Leo Gao,
  Elie Georges, Christian Gibson, Vik Goel, Tarun Gogineni, Gabriel Goh, Rapha Gontijo-Lopes, Jonathan Gordon, Morgan Grafstein, Scott Gray, Ryan Greene, Joshua Gross, Shixiang~Shane Gu, Yufei Guo, Chris Hallacy, Jesse Han, Jeff Harris, Yuchen He, Mike Heaton, Johannes Heidecke, Chris Hesse, Alan Hickey, Wade Hickey, Peter Hoeschele, Brandon Houghton, Kenny Hsu, Shengli Hu, Xin Hu, Joost Huizinga, Shantanu Jain, Shawn Jain, Joanne Jang, Angela Jiang, Roger Jiang, Haozhun Jin, Denny Jin, Shino Jomoto, Billie Jonn, Heewoo Jun, Tomer Kaftan, Łukasz Kaiser, Ali Kamali, Ingmar Kanitscheider, Nitish~Shirish Keskar, Tabarak Khan, Logan Kilpatrick, Jong~Wook Kim, Christina Kim, Yongjik Kim, Hendrik Kirchner, Jamie Kiros, Matt Knight, Daniel Kokotajlo, Łukasz Kondraciuk, Andrew Kondrich, Aris Konstantinidis, Kyle Kosic, Gretchen Krueger, Vishal Kuo, Michael Lampe, Ikai Lan, Teddy Lee, Jan Leike, Jade Leung, Daniel Levy, Chak~Ming Li, Rachel Lim, Molly Lin, Stephanie Lin, Mateusz Litwin, Theresa Lopez, Ryan Lowe,
  Patricia Lue, Anna Makanju, Kim Malfacini, Sam Manning, Todor Markov, Yaniv Markovski, Bianca Martin, Katie Mayer, Andrew Mayne, Bob McGrew, Scott~Mayer McKinney, Christine McLeavey, Paul McMillan, Jake McNeil, David Medina, Aalok Mehta, Jacob Menick, Luke Metz, Andrey Mishchenko, Pamela Mishkin, Vinnie Monaco, Evan Morikawa, Daniel Mossing, Tong Mu, Mira Murati, Oleg Murk, David Mély, Ashvin Nair, Reiichiro Nakano, Rajeev Nayak, Arvind Neelakantan, Richard Ngo, Hyeonwoo Noh, Long Ouyang, Cullen O'Keefe, Jakub Pachocki, Alex Paino, Joe Palermo, Ashley Pantuliano, Giambattista Parascandolo, Joel Parish, Emy Parparita, Alex Passos, Mikhail Pavlov, Andrew Peng, Adam Perelman, Filipe de~Avila Belbute~Peres, Michael Petrov, Henrique~Ponde de~Oliveira~Pinto, Michael, Pokorny, Michelle Pokrass, Vitchyr Pong, Tolly Powell, Alethea Power, Boris Power, Elizabeth Proehl, Raul Puri, Alec Radford, Jack Rae, Aditya Ramesh, Cameron Raymond, Francis Real, Kendra Rimbach, Carl Ross, Bob Rotsted, Henri Roussez, Nick Ryder,
  Mario Saltarelli, Ted Sanders, Shibani Santurkar, Girish Sastry, Heather Schmidt, David Schnurr, John Schulman, Daniel Selsam, Kyla Sheppard, Toki Sherbakov, Jessica Shieh, Sarah Shoker, Pranav Shyam, Szymon Sidor, Eric Sigler, Maddie Simens, Jordan Sitkin, Katarina Slama, Ian Sohl, Benjamin Sokolowsky, Yang Song, Natalie Staudacher, Felipe~Petroski Such, Natalie Summers, Ilya Sutskever, Jie Tang, Nikolas Tezak, Madeleine Thompson, Phil Tillet, Amin Tootoonchian, Elizabeth Tseng, Preston Tuggle, Nick Turley, Jerry Tworek, Juan Felipe~Cerón Uribe, Andrea Vallone, Arun Vijayvergiya, Chelsea Voss, Carroll Wainwright, Justin~Jay Wang, Alvin Wang, Ben Wang, Jonathan Ward, Jason Wei, CJ~Weinmann, Akila Welihinda, Peter Welinder, Jiayi Weng, Lilian Weng, Matt Wiethoff, Dave Willner, Clemens Winter, Samuel Wolrich, Hannah Wong, Lauren Workman, Sherwin Wu, Jeff Wu, Michael Wu, Kai Xiao, Tao Xu, Sarah Yoo, Kevin Yu, Qiming Yuan, Wojciech Zaremba, Rowan Zellers, Chong Zhang, Marvin Zhang, Shengjia Zhao, Tianhao
  Zheng, Juntang Zhuang, William Zhuk, and Barret Zoph. 2023.
\newblock \href {http://arxiv.org/abs/2303.08774} {{GPT-4} technical report}.

\bibitem[{Otegi et~al.(2022)Otegi, {San Vicente}, Saralegi, Pe{\~n}as, Lozano, and Agirre}]{OTEGI2022108072}
Arantxa Otegi, I{\~n}aki {San Vicente}, Xabier Saralegi, Anselmo Pe{\~n}as, Borja Lozano, and Eneko Agirre. 2022.
\newblock \href {https://doi.org/https://doi.org/10.1016/j.knosys.2021.108072} {Information retrieval and question answering: A case study on {COVID-19} scientific literature}.
\newblock \emph{Knowledge-Based Systems}, 240:108072.

\bibitem[{Ouyang et~al.(2022)Ouyang, Wu, Jiang, Almeida, Wainwright, Mishkin, Zhang, Agarwal, Slama, and Ray}]{Ouyang2022}
Long Ouyang, Jeffrey Wu, Xu~Jiang, Diogo Almeida, Carroll Wainwright, Pamela Mishkin, Chong Zhang, Sandhini Agarwal, Katarina Slama, and Alex Ray. 2022.
\newblock {Training language models to follow instructions with human feedback}.
\newblock \emph{Advances in Neural Information Processing Systems}, 35:27730--27744.

\bibitem[{Pandya and Bhatt(2021)}]{pandya2021question}
Hariom~A. Pandya and Brijesh~S. Bhatt. 2021.
\newblock \href {http://arxiv.org/abs/2112.03572} {Question answering survey: Directions, challenges, datasets, evaluation matrices}.

\bibitem[{Papineni et~al.(2002)Papineni, Roukos, Ward, and Zhu}]{papineni-etal-2002-bleu}
Kishore Papineni, Salim Roukos, Todd Ward, and Wei-Jing Zhu. 2002.
\newblock \href {https://doi.org/10.3115/1073083.1073135} {{B}leu: A method for automatic evaluation of machine translation}.
\newblock In \emph{Proceedings of the 40th Annual Meeting of the Association for Computational Linguistics}, pages 311--318, Philadelphia, Pennsylvania, USA. Association for Computational Linguistics.

\bibitem[{Pei et~al.(2020)Pei, Yin, Hawk, Alman, and Gao}]{Pei2020}
Zongrui Pei, Junqi Yin, Jeffrey~A Hawk, David~E Alman, and Michael~C Gao. 2020.
\newblock {Machine-learning informed prediction of high-entropy solid solution formation: Beyond the Hume-Rothery rules}.
\newblock \emph{npj Computational Materials}, 6(1):50.

\bibitem[{Pillutla et~al.(2021)Pillutla, Swayamdipta, Zellers, Thickstun, Welleck, Choi, and Harchaoui}]{pillutla2021mauve}
Krishna Pillutla, Swabha Swayamdipta, Rowan Zellers, John Thickstun, Sean Welleck, Yejin Choi, and Zaid Harchaoui. 2021.
\newblock \href {https://openreview.net/forum?id=Tqx7nJp7PR} {{MAUVE}: Measuring the gap between neural text and human text using divergence frontiers}.
\newblock In \emph{Advances in Neural Information Processing Systems}.

\bibitem[{Rajpurkar et~al.(2016)Rajpurkar, Zhang, Lopyrev, and Liang}]{rajpurkar-etal-2016-squad}
Pranav Rajpurkar, Jian Zhang, Konstantin Lopyrev, and Percy Liang. 2016.
\newblock \href {https://doi.org/10.18653/v1/D16-1264} {{SQ}u{AD}: 100,000+ questions for machine comprehension of text}.
\newblock In \emph{Proceedings of the 2016 Conference on Empirical Methods in Natural Language Processing}, pages 2383--2392, Austin, Texas. Association for Computational Linguistics.

\bibitem[{Rasley et~al.(2020)Rasley, Rajbhandari, Ruwase, and He}]{rasley2020deepspeed}
Jeff Rasley, Samyam Rajbhandari, Olatunji Ruwase, and Yuxiong He. 2020.
\newblock \href {https://doi.org/10.1145/3394486.3406703} {Deep{S}peed: System optimizations enable training deep learning models with over 100 billion parameters}.
\newblock In \emph{Proceedings of the 26th ACM SIGKDD International Conference on Knowledge Discovery \& Data Mining}, pages 3505--3506.

\bibitem[{Rasool et~al.(2024)Rasool, Kurniawan, Balugo, Barnett, Vasa, Chesser, Hampstead, Belleville, Mouzakis, and Bahar-Fuchs}]{rasool2024evaluating}
Zafaryab Rasool, Stefanus Kurniawan, Sherwin Balugo, Scott Barnett, Rajesh Vasa, Courtney Chesser, Benjamin~M. Hampstead, Sylvie Belleville, Kon Mouzakis, and Alex Bahar-Fuchs. 2024.
\newblock \href {http://arxiv.org/abs/2311.07878} {Evaluating {LLMs} on document-based {QA}: Exact answer selection and numerical extraction using cogtale dataset}.

\bibitem[{Roberts et~al.(2020)Roberts, Raffel, and Shazeer}]{roberts2020much}
Adam Roberts, Colin Raffel, and Noam Shazeer. 2020.
\newblock How much knowledge can you pack into the parameters of a language model?
\newblock \emph{arXiv preprint arXiv:2002.08910}.

\bibitem[{Sahu et~al.(2022)Sahu, Rodriguez, Laradji, Atighehchian, Vazquez, and Bahdanau}]{sahu-etal-2022-data}
Gaurav Sahu, Pau Rodriguez, Issam Laradji, Parmida Atighehchian, David Vazquez, and Dzmitry Bahdanau. 2022.
\newblock \href {https://doi.org/10.18653/v1/2022.nlp4convai-1.5} {Data augmentation for intent classification with off-the-shelf large language models}.
\newblock In \emph{Proceedings of the 4th Workshop on NLP for Conversational AI}, pages 47--57, Dublin, Ireland. Association for Computational Linguistics.

\bibitem[{Saikh et~al.(2022)Saikh, Ghosal, Mittal, Ekbal, and Bhattacharyya}]{saikh_scienceqa_2022}
Tanik Saikh, Tirthankar Ghosal, Amish Mittal, Asif Ekbal, and Pushpak Bhattacharyya. 2022.
\newblock \href {https://doi.org/10.1007/s00799-022-00329-y} {{ScienceQA}: A novel resource for question answering on scholarly articles}.
\newblock \emph{International Journal on Digital Libraries}, 23(3):289--301.

\bibitem[{Schwenk et~al.(2022)Schwenk, Khandelwal, Clark, Marino, and Mottaghi}]{schwenk2022okvqa}
Dustin Schwenk, Apoorv Khandelwal, Christopher Clark, Kenneth Marino, and Roozbeh Mottaghi. 2022.
\newblock A-okvqa: A benchmark for visual question answering using world knowledge.
\newblock In \emph{European Conference on Computer Vision}, pages 146--162. Springer.

\bibitem[{Singhal et~al.(2022)Singhal, Azizi, Tu, Mahdavi, Wei, Chung, Scales, Tanwani, Cole-Lewis, Pfohl et~al.}]{singhal2022large}
Karan Singhal, Shekoofeh Azizi, Tao Tu, S~Sara Mahdavi, Jason Wei, Hyung~Won Chung, Nathan Scales, Ajay Tanwani, Heather Cole-Lewis, Stephen Pfohl, et~al. 2022.
\newblock Large language models encode clinical knowledge.
\newblock \emph{arXiv preprint arXiv:2212.13138}.

\bibitem[{Song et~al.(2023)Song, Miret, Zhang, and Liu}]{song-etal-2023-honeybee}
Yu~Song, Santiago Miret, Huan Zhang, and Bang Liu. 2023.
\newblock \href {https://doi.org/10.18653/v1/2023.findings-emnlp.380} {{H}oney{B}ee: Progressive instruction finetuning of large language models for materials science}.
\newblock In \emph{Findings of the Association for Computational Linguistics: EMNLP 2023}, pages 5724--5739, Singapore. Association for Computational Linguistics.

\bibitem[{Subramanian et~al.(2024)Subramanian, Schlegel, Kashyap, Nguyen, Dwivedi, and Winkler}]{subramanian2024mqalm}
Anand Subramanian, Viktor Schlegel, Abhinav~Ramesh Kashyap, Thanh-Tung Nguyen, Vijay~Prakash Dwivedi, and Stefan Winkler. 2024.
\newblock \href {http://arxiv.org/abs/2406.03699} {M-qalm: A benchmark to assess clinical reading comprehension and knowledge recall in large language models via question answering}.

\bibitem[{Tang et~al.(2023)Tang, Lu, Jiang, Huang, Zhang, Zhao, and Wei}]{tang2023metrics}
Tianyi Tang, Hongyuan Lu, Yuchen~Eleanor Jiang, Haoyang Huang, Dongdong Zhang, Wayne~Xin Zhao, and Furu Wei. 2023.
\newblock \href {http://arxiv.org/abs/2305.15067} {Not all metrics are guilty: Improving {NLG} evaluation with {LLM} paraphrasing}.

\bibitem[{Taori et~al.(2023)Taori, Gulrajani, Zhang, Dubois, Li, Guestrin, Liang, and Hashimoto}]{alpaca}
Rohan Taori, Ishaan Gulrajani, Tianyi Zhang, Yann Dubois, Xuechen Li, Carlos Guestrin, Percy Liang, and Tatsunori~B. Hashimoto. 2023.
\newblock Stanford alpaca: An instruction-following llama model.
\newblock \url{https://github.com/tatsu-lab/stanford_alpaca}.

\bibitem[{Taylor et~al.(2022)Taylor, Kardas, Cucurull, Scialom, Hartshorn, Saravia, Poulton, Kerkez, and Stojnic}]{taylor2022galactica}
Ross Taylor, Marcin Kardas, Guillem Cucurull, Thomas Scialom, Anthony Hartshorn, Elvis Saravia, Andrew Poulton, Viktor Kerkez, and Robert Stojnic. 2022.
\newblock Galactica: A large language model for science.
\newblock \emph{arXiv preprint arXiv:2211.09085}.

\bibitem[{Touvron et~al.(2023{\natexlab{a}})Touvron, Lavril, Izacard, Martinet, Lachaux, Lacroix, Rozi{\`e}re, Goyal, Hambro, Azhar et~al.}]{touvron2023llama1}
Hugo Touvron, Thibaut Lavril, Gautier Izacard, Xavier Martinet, Marie-Anne Lachaux, Timoth{\'e}e Lacroix, Baptiste Rozi{\`e}re, Naman Goyal, Eric Hambro, Faisal Azhar, et~al. 2023{\natexlab{a}}.
\newblock Llama: Open and efficient foundation language models.
\newblock \emph{arXiv preprint arXiv:2302.13971}.

\bibitem[{Touvron et~al.(2023{\natexlab{b}})Touvron, Martin, Stone, Albert, Almahairi, Babaei, Bashlykov, Batra, Bhargava, Bhosale, Bikel, Blecher, Ferrer, Chen, Cucurull, Esiobu, Fernandes, Fu, Fu, Fuller, Gao, Goswami, Goyal, Hartshorn, Hosseini, Hou, Inan, Kardas, Kerkez, Khabsa, Kloumann, Korenev, Koura, Lachaux, Lavril, Lee, Liskovich, Lu, Mao, Martinet, Mihaylov, Mishra, Molybog, Nie, Poulton, Reizenstein, Rungta, Saladi, Schelten, Silva, Smith, Subramanian, Tan, Tang, Taylor, Williams, Kuan, Xu, Yan, Zarov, Zhang, Fan, Kambadur, Narang, Rodriguez, Stojnic, Edunov, and Scialom}]{touvron2023llama}
Hugo Touvron, Louis Martin, Kevin Stone, Peter Albert, Amjad Almahairi, Yasmine Babaei, Nikolay Bashlykov, Soumya Batra, Prajjwal Bhargava, Shruti Bhosale, Dan Bikel, Lukas Blecher, Cristian~Canton Ferrer, Moya Chen, Guillem Cucurull, David Esiobu, Jude Fernandes, Jeremy Fu, Wenyin Fu, Brian Fuller, Cynthia Gao, Vedanuj Goswami, Naman Goyal, Anthony Hartshorn, Saghar Hosseini, Rui Hou, Hakan Inan, Marcin Kardas, Viktor Kerkez, Madian Khabsa, Isabel Kloumann, Artem Korenev, Punit~Singh Koura, Marie-Anne Lachaux, Thibaut Lavril, Jenya Lee, Diana Liskovich, Yinghai Lu, Yuning Mao, Xavier Martinet, Todor Mihaylov, Pushkar Mishra, Igor Molybog, Yixin Nie, Andrew Poulton, Jeremy Reizenstein, Rashi Rungta, Kalyan Saladi, Alan Schelten, Ruan Silva, Eric~Michael Smith, Ranjan Subramanian, Xiaoqing~Ellen Tan, Binh Tang, Ross Taylor, Adina Williams, Jian~Xiang Kuan, Puxin Xu, Zheng Yan, Iliyan Zarov, Yuchen Zhang, Angela Fan, Melanie Kambadur, Sharan Narang, Aurelien Rodriguez, Robert Stojnic, Sergey Edunov, and Thomas
  Scialom. 2023{\natexlab{b}}.
\newblock \href {http://arxiv.org/abs/2307.09288} {Llama 2: Open foundation and fine-tuned chat models}.

\bibitem[{Virani et~al.(2023)Virani, Yadav, Sonawane, and Jawale}]{virani2023automatic}
Altaj Virani, Rakesh Yadav, Prachi Sonawane, and Smita Jawale. 2023.
\newblock \href {https://doi.org/10.1109/ICSCSS57650.2023.10169726} {Automatic question answer generation using {T5 and NLP}}.
\newblock In \emph{International Conference on Sustainable Computing and Smart Systems}, pages 1667--1673.

\bibitem[{Wang et~al.(2024)Wang, Yang, Huang, Yang, Majumder, and Wei}]{wang2024improving}
Liang Wang, Nan Yang, Xiaolong Huang, Linjun Yang, Rangan Majumder, and Furu Wei. 2024.
\newblock \href {http://arxiv.org/abs/2401.00368} {Improving text embeddings with large language models}.

\bibitem[{Wang et~al.(2022)Wang, Kordi, Mishra, Liu, Smith, Khashabi, and Hajishirzi}]{wang2022self}
Yizhong Wang, Yeganeh Kordi, Swaroop Mishra, Alisa Liu, Noah~A Smith, Daniel Khashabi, and Hannaneh Hajishirzi. 2022.
\newblock \href {https://arxiv.org/abs/2212.10560} {Self-{I}nstruct: Aligning language model with self generated instructions}.
\newblock \emph{arXiv preprint arXiv:2212.10560}.

\bibitem[{Wang et~al.(2023)Wang, Kordi, Mishra, Liu, Smith, Khashabi, and Hajishirzi}]{wang-etal-2023-self-instruct}
Yizhong Wang, Yeganeh Kordi, Swaroop Mishra, Alisa Liu, Noah~A. Smith, Daniel Khashabi, and Hannaneh Hajishirzi. 2023.
\newblock \href {https://doi.org/10.18653/v1/2023.acl-long.754} {Self-{I}nstruct: Aligning language models with self-generated instructions}.
\newblock In \emph{Proceedings of the 61st Annual Meeting of the Association for Computational Linguistics (Volume 1: Long Papers)}, pages 13484--13508, Toronto, Canada. Association for Computational Linguistics.

\bibitem[{Wei et~al.(2022)Wei, Wang, Schuurmans, Bosma, Ichter, Xia, Chi, Le, and Zhou}]{NEURIPS2022_9d560961}
Jason Wei, Xuezhi Wang, Dale Schuurmans, Maarten Bosma, Brian Ichter, Fei Xia, Ed~Chi, Quoc~V Le, and Denny Zhou. 2022.
\newblock \href {https://proceedings.neurips.cc/paper_files/paper/2022/file/9d5609613524ecf4f15af0f7b31abca4-Paper-Conference.pdf} {Chain-of-thought prompting elicits reasoning in large language models}.
\newblock In \emph{Advances in Neural Information Processing Systems}, volume~35, pages 24824--24837. Curran Associates, Inc.

\bibitem[{Welbl et~al.(2017)Welbl, Liu, and Gardner}]{welbl-etal-2017-crowdsourcing}
Johannes Welbl, Nelson~F. Liu, and Matt Gardner. 2017.
\newblock \href {https://doi.org/10.18653/v1/W17-4413} {Crowdsourcing multiple choice science questions}.
\newblock In \emph{Proceedings of the 3rd Workshop on Noisy User-generated Text}, pages 94--106, Copenhagen, Denmark. Association for Computational Linguistics.

\bibitem[{Xie et~al.(2023{\natexlab{a}})Xie, Wan, Huang, Yin, Liu, Wang, Linghu, Kit, Grazian, Zhang, Razzak, and Hoex}]{xie2023darwin}
Tong Xie, Yuwei Wan, Wei Huang, Zhenyu Yin, Yixuan Liu, Shaozhou Wang, Qingyuan Linghu, Chunyu Kit, Clara Grazian, Wenjie Zhang, Imran Razzak, and Bram Hoex. 2023{\natexlab{a}}.
\newblock \href {http://arxiv.org/abs/2308.13565} {{DARWIN} series: Domain specific large language models for natural science}.

\bibitem[{Xie et~al.(2023{\natexlab{b}})Xie, Wan, Huang, Zhou, Liu, Linghu, Wang, Kit, Grazian, Zhang et~al.}]{xie2023large}
Tong Xie, Yuwei Wan, Wei Huang, Yufei Zhou, Yixuan Liu, Qingyuan Linghu, Shaozhou Wang, Chunyu Kit, Clara Grazian, Wenjie Zhang, et~al. 2023{\natexlab{b}}.
\newblock Large language models as master key: unlocking the secrets of materials science with gpt.
\newblock \emph{arXiv preprint arXiv:2304.02213}.

\bibitem[{Xie et~al.(2023{\natexlab{c}})Xie, Wan, Lu, Zhang, Kit, and Hoex}]{xie2023tokenizer}
Tong Xie, Yuwei Wan, Ke~Lu, Wenjie Zhang, Chunyu Kit, and Bram Hoex. 2023{\natexlab{c}}.
\newblock Tokenizer effect on functional material prediction: Investigating contextual word embeddings for knowledge discovery.
\newblock In \emph{AI for Accelerated Materials Design-NeurIPS 2023 Workshop}.

\bibitem[{Xie et~al.(2024)Xie, Wan, Zhou, Huang, Liu, Linghu, Wang, Kit, Grazian, Zhang et~al.}]{xie2024creation}
Tong Xie, Yuwei Wan, Yufei Zhou, Wei Huang, Yixuan Liu, Qingyuan Linghu, Shaozhou Wang, Chunyu Kit, Clara Grazian, Wenjie Zhang, et~al. 2024.
\newblock Creation of a structured solar cell material dataset and performance prediction using large language models.
\newblock \emph{Patterns}, 5(5).

\bibitem[{Xu et~al.(2024)Xu, Jain, and Kankanhalli}]{xu2024hallucination}
Ziwei Xu, Sanjay Jain, and Mohan Kankanhalli. 2024.
\newblock Hallucination is inevitable: An innate limitation of large language models.
\newblock \emph{arXiv preprint arXiv:2401.11817}.

\bibitem[{Ye et~al.(2024)Ye, Ren, Wang, Wan, Razzak, Xie, and Zhang}]{ye2024construction}
Yanpeng Ye, Jie Ren, Shaozhou Wang, Yuwei Wan, Imran Razzak, Tong Xie, and Wenjie Zhang. 2024.
\newblock Construction of functional materials knowledge graph in multidisciplinary materials science via large language model.
\newblock \emph{arXiv preprint arXiv:2404.03080}.

\bibitem[{Yoo et~al.(2021)Yoo, Park, Kang, Lee, and Park}]{yoo-etal-2021-gpt3mix-leveraging}
Kang~Min Yoo, Dongju Park, Jaewook Kang, Sang-Woo Lee, and Woomyoung Park. 2021.
\newblock \href {https://doi.org/10.18653/v1/2021.findings-emnlp.192} {{GPT}3{M}ix: Leveraging large-scale language models for text augmentation}.
\newblock In \emph{Findings of the Association for Computational Linguistics: EMNLP 2021}, pages 2225--2239, Punta Cana, Dominican Republic. Association for Computational Linguistics.

\bibitem[{Zhang et~al.(2024{\natexlab{a}})Zhang, Liu, Tan, Chen, Yan, Yan, Li, Huang, Yue, Zhou et~al.}]{zhang2024chemllm}
Di~Zhang, Wei Liu, Qian Tan, Jingdan Chen, Hang Yan, Yuliang Yan, Jiatong Li, Weiran Huang, Xiangyu Yue, Dongzhan Zhou, et~al. 2024{\natexlab{a}}.
\newblock Chemllm: A chemical large language model.
\newblock \emph{arXiv preprint arXiv:2402.06852}.

\bibitem[{Zhang et~al.(2024{\natexlab{b}})Zhang, Ding, Lyv, Wang, Yin, Zhang, Yu, Wang, Li, Xiang et~al.}]{zhang2024scientific}
Qiang Zhang, Keyang Ding, Tianwen Lyv, Xinda Wang, Qingyu Yin, Yiwen Zhang, Jing Yu, Yuhao Wang, Xiaotong Li, Zhuoyi Xiang, et~al. 2024{\natexlab{b}}.
\newblock Scientific large language models: A survey on biological \& chemical domains.
\newblock \emph{arXiv preprint arXiv:2401.14656}.

\bibitem[{Zhang et~al.(2020)Zhang, Kishore, Wu, Weinberger, and Artzi}]{Zhang2020BERTScore}
Tianyi Zhang, Varsha Kishore, Felix Wu, Kilian~Q. Weinberger, and Yoav Artzi. 2020.
\newblock \href {https://openreview.net/forum?id=SkeHuCVFDr} {{BERTScore}: Evaluating text generation with {BERT}}.
\newblock In \emph{International Conference on Learning Representations}.

\bibitem[{Zheng et~al.(2023)Zheng, Chiang, Sheng, Zhuang, Wu, Zhuang, Lin, Li, Li, Xing, Zhang, Gonzalez, and Stoica}]{zheng2023judging}
Lianmin Zheng, Wei-Lin Chiang, Ying Sheng, Siyuan Zhuang, Zhanghao Wu, Yonghao Zhuang, Zi~Lin, Zhuohan Li, Dacheng Li, Eric Xing, Hao Zhang, Joseph~E. Gonzalez, and Ion Stoica. 2023.
\newblock \href {https://openreview.net/forum?id=uccHPGDlao} {Judging {LLM}-as-a-judge with {MT}-bench and {Chatbot Arena}}.
\newblock In \emph{Thirty-seventh Conference on Neural Information Processing Systems Datasets and Benchmarks Track}.

\bibitem[{Zhuo et~al.(2018)Zhuo, {Mansouri Tehrani}, and Brgoch}]{Zhuo2018}
Ya~Zhuo, Aria {Mansouri Tehrani}, and Jakoah Brgoch. 2018.
\newblock {Predicting the band gaps of inorganic solids by machine learning}.
\newblock \emph{The journal of physical chemistry letters}, 9(7):1668--1673.

\end{thebibliography}

\appendix

\section{Complete list of 24 WoS categories and original distribution}
\label{sec:Appendix A}
\begin{enumerate}[itemsep=-1mm]
\item Materials Science, Multidisciplinary
\item Physics, Multidisciplinary
\item Chemistry, Multidisciplinary
\item Materials Science, Coatings \& Films
\item Materials Science, Ceramics
\item Materials Science, Characterization \& Testing
\item Materials Science, Biomaterials
\item Materials Science, Composites
\item Materials Science, Paper \& Wood 
\item Materials Science, Textiles
\item Physics, Applied
\item Physics, Atomic, Molecular \& Chemical
\item Physics, Condensed Matter
\item Physics, Fluids \& Plasmas
\item Physics, Mathematical
\item Physics, Nuclear
\item Physics, Particles \& Fields
\item Chemistry, Analytical
\item Chemistry, Applied
\item Chemistry, Inorganic \& Nuclear
\item Chemistry, Medicinal
\item Chemistry, Organic
\item Chemistry, Physical
\item Energy \& Fuels
\end{enumerate}

\begin{figure*}
\centering
  \includegraphics[width=0.85\linewidth]{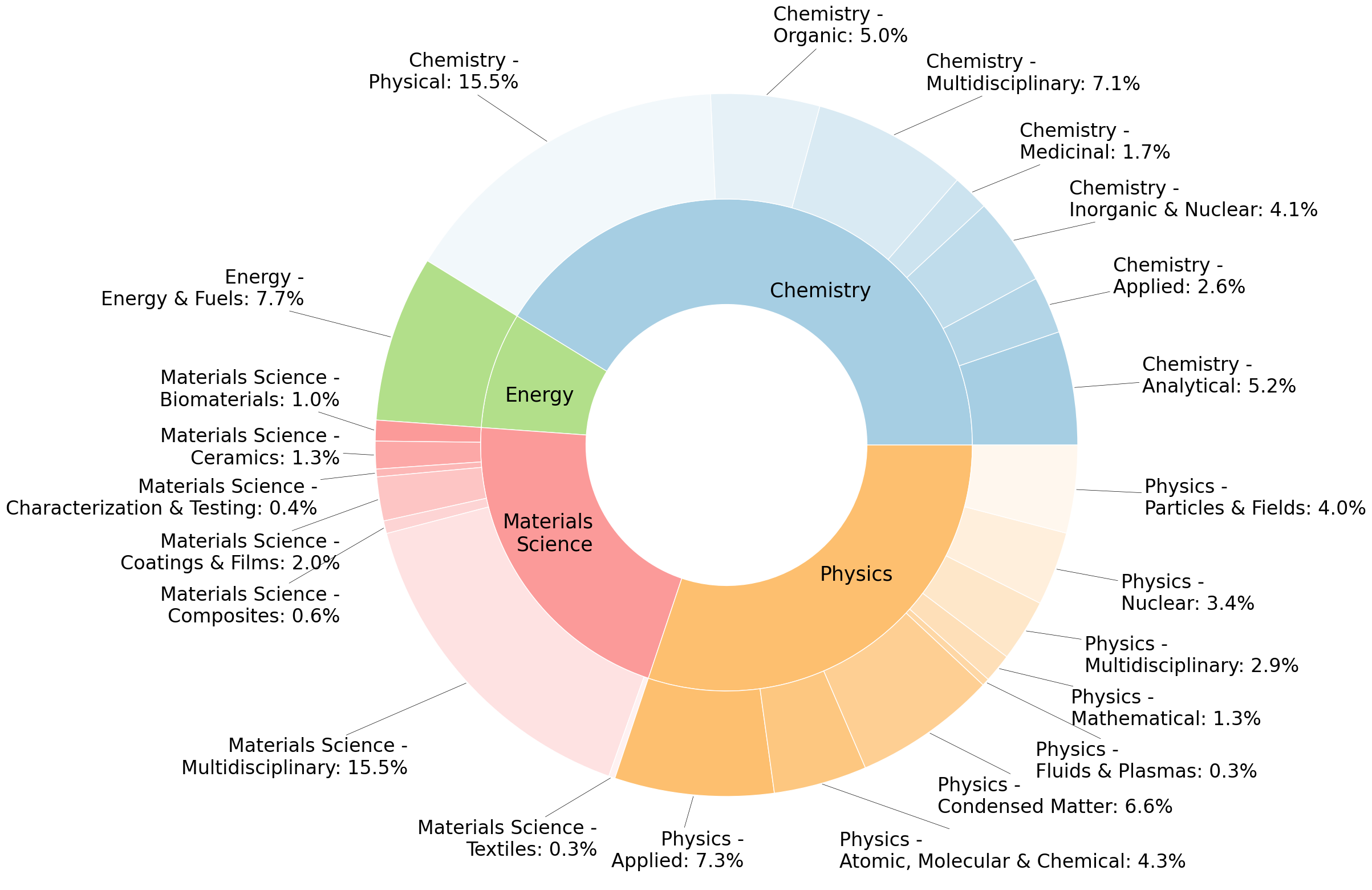}
  \caption{Original distribution of papers from the WoS Core Collection across 24 WoS categories selected from Chemistry, Physics, Materials Science and Energy.}
\label{fig:4}
\end{figure*}

\autoref{fig:4} shows the original distribution of 24 categories.

\section{Statistics of SciQAG-24D dataset}
\label{sec: statistics}

\autoref{table:2} shows the details of 24 categories in SciQAG-24D dataset.

\begin{table*}[ht!]
\centering
\begin{adjustbox}{width=1\textwidth}
\begin{tabular}{l|c|cc|cc|cc}
\hline
\multirow{2}{1pt}{Category} & Original & \multicolumn{4}{c|}{Support (train)} & \multicolumn{2}{c}{Support (test)} \\
\cline{2-8}
&  \begin{tabular}[c]{@{}r@{}}Paper\end{tabular} & \begin{tabular}[c]{@{}r@{}}Paper\end{tabular} & \begin{tabular}[c]{@{}r@{}}QA\end{tabular} & \begin{tabular}[c]{@{}r@{}}Paper\_filter\end{tabular} & \begin{tabular}[c]{@{}r@{}}QA\_filter\end{tabular} & \begin{tabular}[c]{@{}r@{}}Paper\_filter\end{tabular} & \begin{tabular}[c]{@{}r@{}}QA\_filter\end{tabular} \\
\cline{1-8}
Chemistry, Analytical & 1020 & 970 & 9700 & 970 & 8075 &  50 & 375 \\
Chemistry, Applied & 1037 & 987 & 9870 & 987 & 8314 &  50 & 385 \\
Physics, Condensed Matter & 1025 & 975 & 9750 & 974 & 8340 &  50 & 366 \\
Chemistry, Organic & 1048 & 998 & 9980 & 998 & 8513 &  50 & 369 \\
Materials Science, Multidisciplinary & 1000 & 950 & 9500 & 950 & 8382 &  50 & 378 \\
Chemistry, Inorganic \& Nuclear & 1024 & 974 & 9740 & 973 & 8225 &  50 & 333 \\
Physics, Nuclear & 941 & 891 & 8910 & 888 & 7208 &  50 & 374 \\
Physics, Atomic, Molecular \& Chemical & 1081 & 1031 & 10310 & 1031 & 8215 &  50 & 324 \\
Materials Science, Paper \& Wood & 375 & 325 & 3250 & 325 & 2743 &  50 & 319 \\
Physics, Mathematical & 1036 & 986 & 9860 & 985 & 7126 &  50 & 257 \\
Physics, Particles \& Fields & 1000 & 950 & 9500 & 944 & 7361 &  50 & 341 \\
Materials Science, Biomaterials & 1007 & 957 & 9570 & 957 & 8093 &  50 & 357 \\
Physics, Multidisciplinary & 1000 & 950 & 9500 & 950 & 8009 &  50 & 358 \\
Physics, Applied & 581 & 531 & 5310 & 531 & 4699 &  50 & 391 \\
Energy \& Fuels & 1076 & 1026 & 10260 & 1026 & 8625 &  50 & 390 \\
Materials Science, Coatings \& Films & 1004 & 954 & 9540 & 954 & 8390 &  50 & 377 \\
Chemistry, Multidisciplinary & 952 & 902 & 9020 & 901 & 7777 &  50 & 368 \\
Chemistry, Medicinal & 570 & 520 & 5200 & 520 & 4302 &  50 & 335 \\
Materials Science, Textiles & 1023 & 973 & 9730 & 973 & 8095 &  50 & 349 \\
Materials Science, Characterization \& Testing & 1048 & 998 & 9980 & 998 & 8517 &  49 & 367 \\
Physics, Fluids \& Plasmas & 1016 & 966 & 9660 & 965 & 7085 &  50 & 326 \\
Materials Science, Composites & 1046 & 996 & 9960 & 996 & 8541 &  50 & 376 \\
Chemistry, Physical & 791 & 741 & 7410 & 741 & 6380 &  50 & 358 \\
Materials Science, Ceramics & 1042 & 992 & 9920 & 992 & 8496 &  50 & 358 \\
\hline
Sum & 22743 & 21543 & 215430 & 21529 & 179511 & 1199 & 8531 \\
\hline
\end{tabular}
\end{adjustbox}
\caption{Statistics of SciQAG-24D dataset.}
\label{table:2}
\end{table*}

\section{Prompts for QA generator}
\label{sec:Appendix B}
Here is a scientific paper:\\
\{text\}
\\
Here is a two-step task for you. \\
Step 1: Read given scientific paper and extract a list of 15 keywords focusing on the important terms and concepts within the paragraph. Avoid generic or broad words.\\
Step 2: Generate 10 scientific Q\&A pairs as diverse as possible based on facts and knowledge presented in given paper, focusing on keywords you generated. Keep the following requirements in mind: Avoid ask simple or definitional questions. Please assume that there is no corresponding paper to refer to when asking questions, so ensure that the questions and answers are self-contained (do not refer to external sources like figures or tables, or use demonstrative pronouns such as "this"). Incorporate specific data and insights from the paper to create detailed and informative answers. Ensure that the answers are concise, accurate, and directly related to the corresponding questions. \\
Please present the generated keywords and question-answer pairs in the following format:\\
Keywords: [keyword 1], [keyword 2], ..., [keyword15]\\
Q1: [Question 1]\\
A1: [Answer 1]\\
\\
Q2: [Question 2]\\
A2: [Answer 2]\\
\\
...\\
\\
Q10: [Question 10]\\
A10: [Answer 10]\\

\section{GPT-4 prompts for RACAR metric}
\label{sec:Appendix C}
%\begin{itemize}
%    \item 
    \textbf{Relevance}: Given a scientific paper and questions generated from it, evaluate the relevance of the question to the paper and return a score ranging from 1--3 and give reasons as to why this score was assigned. The output must be a list of dictionaries corresponding to each question, with the fields `score' and `reasons.' If the question does not pertain to the paper, assign a score of 1.

    \vspace{0.5ex}
\noindent
    \underline{Paper}: \{ paper\_text \}

    \vspace{0.5ex}
\noindent
    \underline{Questions}: \{ [ \{ questions \} ] \}

    \vspace{0.5ex}
\noindent
    \underline{Output}:
    
    %\item 
    \vspace{1ex}
    \noindent
    \textbf{Agnosticism}: Given questions generated from a scientific article, evaluate its context independence and return a score ranging from 1--3. Identify whether the question is referring to specific experimental setups, figures, or tables from the paper. For example, questions like ``What can we say about the effect of varying pressure from Figure 1?'' should be assigned a score of 1. The output must be a list of dictionaries corresponding to each question, with the fields `score' and `reasons.'

        \vspace{0.5ex}
\noindent
    \underline{Questions}: \{ [ \{ questions \} ] \}

    \vspace{0.5ex}
\noindent
    \underline{Output}:

    %\item 
        \vspace{1ex}
    \noindent
    \textbf{Completeness}: Given a scientific paper and question answer pairs generated from it, evaluate the completeness of the answer for each question and return a score ranging from 1--3 indicating the extent to which the answer fully addresses the question using the information in the paper, including all subquestions. Also give reasons for assigning the score. The output must be a list of dictionaries for each question answer pair, with the fields `score' and `reasons.'

    \vspace{0.5ex}
\noindent
    \underline{Paper}: \{ paper\_text \}

    \vspace{0.5ex}
\noindent
    \underline{Questions}: \{ [ \{ qa\_pairs \} ] \}

    \vspace{0.5ex}
\noindent
    \underline{Output}:
    
    %\item 
        \vspace{1ex}
    \noindent
    \textbf{Accuracy}: Given a scientific paper and question answer pairs generated from this scientific paper, evaluate the accuracy of the answer for each question and return a score ranging from 1--3 indicating whether the answer is accurately extracted from the paper and give reasons as to why this score was assigned. This involves checking the accuracy of any claims or statements made in the text, and verifying that they are supported by evidence. The output must be a list of dictionaries for each question answer pair, with the fields `score' and `reasons.'

    \vspace{0.5ex}
\noindent
    \underline{Paper}: \{ paper\_text \}

    \vspace{0.5ex}
\noindent
    \underline{QA pairs}: \{ [  \{qa\_pairs\} ] \}

    \vspace{0.5ex}
\noindent
    \underline{Output}:
    
    %\item 
        \vspace{1ex}
    \noindent
    \textbf{Reasonableness}: Given a scientific paper and statements, evaluate the reasonableness of the statements with respect to the paper and return a score ranging from 1--3 indicating how logically consistent the content is, with no obvious contradictions and provide reasons for assigning the score. The output must be a list of dictionaries for each statement, with the fields `score' and `reasons.' Assign a score of 1 if the statement has logical error like contradicts.

    \vspace{0.5ex}
\noindent
    \underline{Paper}: \{ paper\_text \}

    \vspace{0.5ex}
\noindent
    \underline{Statements}: \{ [  \{ answers \} ] \}

    \vspace{0.5ex}
\noindent
    \underline{Output}: 
%\end{itemize}

\section{Prompts for similarity of questions}
\label{sec:Appendix D}
"""<s> [INST] **Objective**: Assess the similarity of two questions not just based on their word content and relatedness, but primarily based on their underlying purpose or intent. This assessment should distinguish between questions that are about opposite aspects (even if related) and those that are different in wording but similar in the information they seek. Mark it from 0--100. 0 means absolutely no same intention, 100 means completely same intention. You don't need to explain your choice, just output a score in the format of similarity score: <the score>.\\    
    **Procedure**:\\
    1. Read and Understand the Questions:\\
        - Carefully read the given questions.\\
        - Understand the core purpose behind each question.\\
    2. Identify Key Aspects:\\
        - Determine what each question is specifically asking for.\\
        - Identify if the questions are seeking positive, negative, or neutral aspects of a subject.\\
    3. Analyze Semantic Differences:\\
        - Compare the intent behind the questions.\\
        - Look for opposing themes such as benefits vs drawbacks, advantages vs disadvantages, etc.\\
    4. Similarity Scoring:\\
        - Assign a high similarity score if both questions are asking about the same aspect of a subject, even if the wording is different.\\
        - Assign a low similarity score if the questions are asking about opposite aspects, even if they are related or use similar words.\\
        \\
    **Examples**:\\
    Question 1: ``What are the pros of using zeolites?''\\
    Question 2: ``What are the drawbacks of using zeolites?''\\
    Analysis: Although both questions are about zeolites, they asks different aspects: positive vs.\ negative.\\
    Similarity Score: 0\\
    \\
    Question 1: ``Give some examples of using zeolites as catalysts."\\
    Question 2: ``What are potential applications of using zeolites as catalysts?''\\
    Analysis: Both questions seek information on how zeolites are used as catalysts, with slightly different wording.\\
    Similarity Score: 100\\
    \\
    **Notes:**\\
    Pay attention to nuanced language that may indicate different intentions even in closely related topics.\\
    Remember, a high degree of relatedness does not necessarily imply a high degree of similarity in intent.\\
    \\
    **Questions**\\
    Question1: {sent1}\\
    Question2: {sent2} [/INST]"""\\

\section{Self-questioning vs. SciQAG}
\label{sec:Appendix E}
\textbf{Self-questioning} \cite{fang2024molinstructions}

\vspace{1ex}
\noindent 
\underline{Prompt}:

\vspace{0.5ex}
\noindent
Given a paragraph of paper, can you ask five questions and give answers that can be directed to the paragraph? List in the format of Question-Answer pair.

\vspace{0.5ex}
\noindent
Paragraph: \{paragraph\}

\vspace{1ex}
\noindent
\underline{Example GPT-3.5 completion}:

\vspace{0.5ex}
\noindent
Q1: What is being discussed in this paragraph? \\
A1: This paragraph is discussing the synthesis of one-dimensional carbon nanorods through a self-templated, catalyst-free strategy and the potential application of these nanorods and graphene nanoribbons in supercapacitor electrodes. \\
\\
Q2: What transforms the nanorods into graphene nanoribbons? \\
A2: Sonochemical treatment followed by chemical activation transforms the nanorods into graphene nanoribbons.\\
\\
Q3: What properties of nanomaterials makes them attractive? \\
A3: The extraordinary electrical, mechanical and thermal properties of one- and two-dimensional carbon nanomaterials makes them attractive. \\
\\
Q4: What is the major challenge associated with the synthesis of these materials? \\
A4: The major challenge in the synthesis of these materials is to develop simple and efficient methods that consume low amounts of energy.  \\
\\
Q5: What advantages does this particular synthetic approach offer?\\
A5: This particular synthetic approach offers advantages such as scalability and the production of functionally useful materials on industrial levels.

\vspace{1ex}
\noindent
\textbf{SciQAG} (this paper)

\vspace{1ex}
\noindent
\underline{Prompt}:

\vspace{0.5ex}
\noindent
Here is a scientific paper:\\
\{text\}

\vspace{0.5ex}
\noindent
Given the provided scientific paper, please complete the following two steps:\\
Step 1: Keyword Extraction\\
Read the scientific paper and identify 15 keywords that capture the most important terms and concepts in the paper, avoiding generic or broad terms. Compile the selected keywords into a list.\\
Step 2: Question-Answer Generation\\
Generate 10 science question-answer pairs as diverse as possible based on facts and knowledge presented in the given paper, focusing on keywords you generated. Keep the following requirements in mind: Avoid asking simple or definitional questions. Assume that the reader does not have access to the original paper or any external sources, so ensure that the questions and answers are self-contained and do not rely on references to figures, tables, or other parts of the paper. Incorporate specific data and insights from the paper to provide detailed and informative answers. Keep the answers concise, accurate, and directly related to the corresponding questions.\\
\\
Please present the generated keywords and question-answer pairs in the following format:\\
Keywords: [keyword 1], [keyword 2], ..., [keyword15]\\
\\
Q1: [Question 1]\\
A1: [Answer 1]\\
\\
Q2: [Question 2]\\
A2: [Answer 2]\\
\\
...\\
\\
Q10: [Question 10]\\
A10: [Answer 10]\\

\vspace{0.5ex}
\noindent
\underline{Example SciQAG completion}:

\vspace{0.5ex}
\noindent
Q1: What is Type 1 diabetes and what causes it?\\
A1: Type 1 diabetes is an autoimmune disease that results from the selective destruction of insulin-producing beta cells in the pancreatic islets. Genetics, environmental factors, nutritional effects, and a combination of these are believed to be associated with the disease. Streptozotocin (STZ) is commonly used to induce diabetes mellitus in experimental studies.\\
\\
Q2: What is the role of metabolomics in healthcare?\\
A2: Metabolomics is a rapidly growing field that relates biological end points to multiple altered metabolite concentrations, providing a wealth of biological information on complex systems. It has been applied to a variety of diseases such as cancer, type 2 diabetes, and inborn errors of metabolism. It uses advanced analytical techniques such as nuclear magnetic resonance (NMR) spectroscopy and mass spectrometry with multivariate statistical analysis to identify potential biomarkers and biological networks.\\
\\
Q3: How was the diabetic rat model established in this study?\\
A3: In this study, female rats were divided into a diabetic group and a control group. The diabetic group received an intraperitoneal dose of STZ, and showed glucose levels in the blood of more than 200 mg/dl after 4 days. The control group was matched for age and gender, and did not receive the STZ injection.\\
\\
Q4: What were the key findings regarding glucose metabolism in the study?\\
A4: The study found that in diabetes, glucose excretion was enormous, being nearly 7500-fold higher than in control rats. Other significantly enhanced metabolites included lactate, alanine, ethanol, and fumarate. These findings suggest that glucose metabolism is dramatically perturbed in diabetes.\\
\\
Q5: What is the importance of the Cori cycle in maintaining glucose homeostasis?\\
A5: The Cori cycle, also known as the glucose-lactate cycle, is a metabolic pathway that plays a crucial role in maintaining glucose homeostasis. It involves the conversion of glucose, lactate, and alanine, and is responsible for producing Cori cycle intermediates, which are used in gluconeogenesis precursor production and facilitating glucose oxidation.\\
\\
Q6: What is the role of acetate switch in metabolism?\\
A6: The acetate switch, also known as the molecular switch that regulates the dissimilation and assimilation of acetate, plays a role in accumulating an acetyl–coenzyme A (CoA) pool due to both exogenous (bacterial fermentation) and endogenous (mammalian) production. This accumulation can have significant implications for metabolism.\\
\\
Q7: How did the urinary metabolites differ between diabetic and control rats in the study?\\
A7: The urinary metabolites showed dramatic changes in the rat diabetic model when compared with blood-based assays. Glucose was the most notable metabolite in diabetic urine, with very high-intensity signals. Other metabolites such as alanine, lactate, ethanol, acetate, and fumarate also showed significant increases in the diabetic group.\\
\\
Q8: What was the effect of STZ administration on plasma glucose levels?\\
A8: Following STZ administration, plasma glucose showed elevated levels, with its quantity being nearly three times higher than in control rats.\\
\\
Q9: What were some of the changes observed in the rat diabetic model's blood parameters?\\
A9: Other than the significant increases in blood glucose levels, the study also observed elevated levels of urea and creatinine, and decreased levels of plasma protein and cholesterol in diabetic rats.\\
\\
Q10: How did the correlations among the detected metabolites differ between the diabetic and control rats?\\
A10: The correlations among the detected metabolites showed a number of striking differences between the diabetic and control rats. For example, the correlation of glucose with alanine was increased in diabetic rats, while the correlation of glucose with lactate decreased. These changes indicate significant alterations in metabolism in the diabetic rats.\\
\\
Analysis:\\
Due to the nature of its prompting technique, self-questioning is prone to generate more generic and contextually tied questions (e.g., ``What is being discussed in this paragraph?''). SciQAG is able to avoid self-referential expressions such as `this,' `that,' and `these,' making it more suitable for crafting challenging closed-book QA datasets, though this may slightly compromise its compatibility with source papers for open-book QA applications.

The scope of questions generated by SciQAG is notably broad and deep. Unlike simpler, definition-focused queries, SciQAG explores various aspects such as relevant properties, application impacts, strategic influences, structural differences, challenges, and experimental insights. This diversity ensures that the questions not only span a wide range of topics but also deeply engage with the subject matter of the scientific papers. The answers produced by SciQAG are equally impressive in their informativeness and specificity. They meticulously avoid over-generalization, often incorporating precise details like specific chemical elements and numerical values that are referenced in the original papers. This level of detail in the answers highlights the method’s ability to maintain relevance and accuracy in relation to the scientific content. The comparison between SciQAG and self-questioning thus underscores SciQAG's strengths in generating diverse, context-independent, and content-rich QA pairs.

\section{10 generated questions example}
\label{sec:Appendix F}
The following 10 questions were generated from a random paper \citet{ivleva2009towards} by QA generator.\\
\\
Q1: What are the applications of rapid diagnostic assays as discussed in the paper?\\
\\
Q2: What is the significance of mycotoxin detection in the food and feed sector?\\
\\
Q3: What are the advantages of lateral-flow devices (LFDs) in the context of rapid immunoassay-based tests?\\
\\
Q4: What are the potential applications of LFDs according to the paper?\\
\\
Q5: What are the challenges in the development of rapid test strips for mycotoxins?\\
\\
Q6: What is the significance of antibody sensitivity and specificity in the performance of a membrane-based immunoassay?\\
\\
Q7: What extraction procedures are commonly used for mycotoxin analysis?\\
\\
Q8: How does matrix variation affect the results of rapid tests for mycotoxins?\\
\\
Q9: What is the role of colloidal gold in test strips for mycotoxins?\\
\\
Q10: How do regulatory limits vary for mycotoxins in different countries?\\

\section{Pairwise visualization}
\begin{figure*}[ht!]
\centering
  \begin{subfigure}{0.49\textwidth}
  \centering
    \includegraphics[height=70mm,trim=2mm 0 30mm 0,clip]{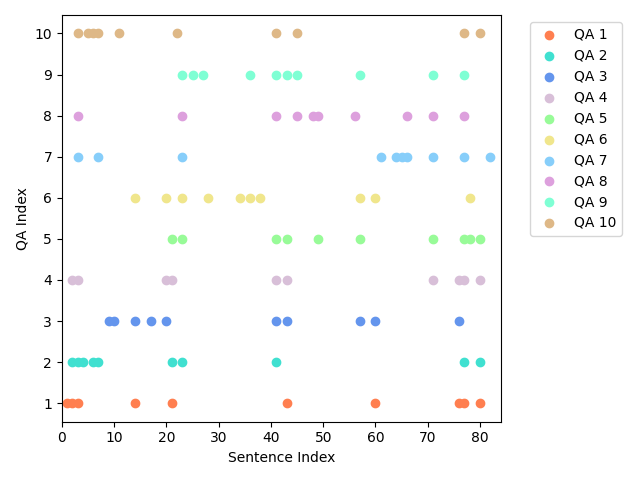}
        
    \vspace{-1ex}
    
    \caption{}
  \end{subfigure}
\hspace{-2ex}
  %\hfill %%
  \begin{subfigure}{0.49\textwidth}
  \centering
    \includegraphics[height=70mm,trim=14mm 0 2mm 0,clip]{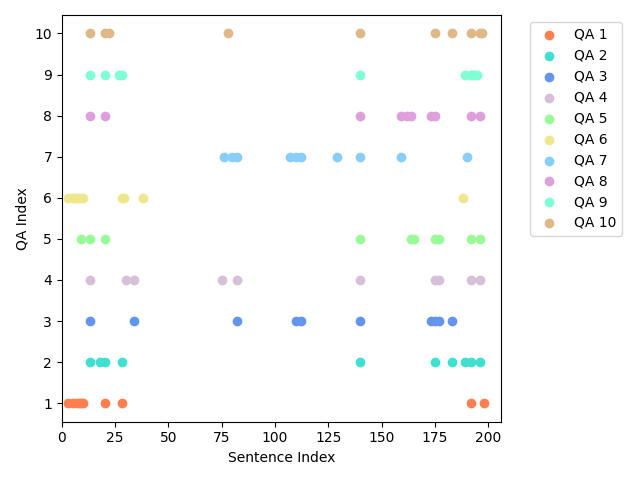}
        
    \vspace{-1ex}
    
    \caption{}
  \end{subfigure}

  \vspace{1ex}
  
    \begin{subfigure}{\textwidth}
    \centering
    \includegraphics[width=\textwidth,trim=0 0 0 9mm,clip]{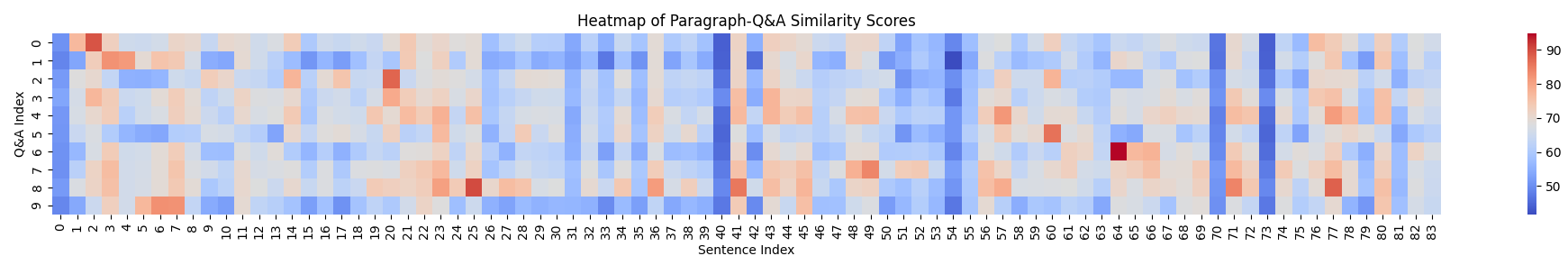}
    
    \vspace{-1ex}
    
    \caption{}
  \end{subfigure}

  \vspace{1ex}

  \begin{subfigure}{\textwidth}
    \centering
    \includegraphics[width=\textwidth,trim=0 0 0 9mm,clip]{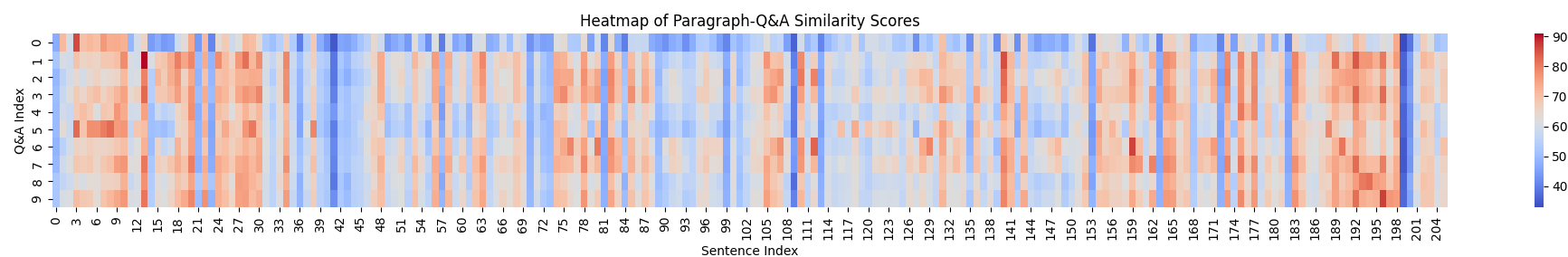}
        
    \vspace{-1ex}
    
    \caption{}
  \end{subfigure}  
  \caption{Pairwise similarity from \citet{krska2009rapid} and \citet{ivleva2009towards}, respectively. 
  (a) and (b): The 10 sentences with the highest similarity scores. 
  (c) and (d): Heatmap of all similarity scores.
  }
  \label{fig:5}
\end{figure*}

We show in \autoref{fig:5} (a) and (b) the top 10 sentences with the highest similarity scores for each answer generated for two articles selected at random from our collection. We observe that some sentences serve as common sources for multiple answers; however, overall, there tend to be distinct sources for each answer. Additionally, we plot heatmaps to provide a holistic view of the similarity scores between each answer and all sentences in the paper. \autoref{fig:5} (c) and (d) offer a thorough overview of the connections between answers and various parts of the given paper (a \& c, b \& d are from the same paper). From the perspective both of coverage and sourcing rates, and of the two visualization methods, it is evident that the answers generated by SciQAG exhibit extensive coverage of the source paper and originate from diverse sources.

\section{Details of generating SciQAG-24D dataset}
We employ the \texttt{gpt-4-1106-preview} model with its default parameter settings for the seed QA generation and RACAR evaluation. We follow the training method from LongLoRA \cite{chen2023longlora} and fine-tune the \texttt{vicuna-7b-v1.5-16k} model \cite{zheng2023judging} using Flash-Attention2 \cite{dao2023flashattention2} and DeepSpeed \cite{rasley2020deepspeed} stage 3 on an 8×A800 GPU machine for 5 epochs with a per-device batch size of 1 and max length of 16,000 for both training and evaluation. Gradients were accumulated over 2 steps and the learning rate was set to 2e-5 with no weight decay. The warm-up steps were set to 20. For inference, we set temperature to 0.8 and top\_p to 0.75, which makes text generated more logical with rich vocabulary. 

\section{Prompts for CAR metric}
\label{sec:Appendix I}
For this task, you are provided with a question-answer pair. Evaluate the quality of answer on the following three criteria and record your evaluations in a score ranging from 1 to 5 for each criterion and provide reasons for assigning the score:\\
1. **Accuracy**: Score the accuracy of the answer in relation to the question. A score of 5 means the answer is fully accurate. This involves checking the accuracy of any claims or statements made in the text, and verifying that they are supported by evidence. While a score of 1 indicates significant inaccuracies.\\
2. **Completeness**: Rate how comprehensive the answer is. A score of 5 indicates that the answer addresses all key points of the question and includes sufficient background and supporting details and evidence. A score of 1 means the answer is largely incomplete.\\
3. **Reasonableness**: Evaluate the logical consistency and reasonableness of the answer. A score of 5 indicates that the answer is logically sound with no contradictions; a score of 1 indicates that the answer contains major contradictions.\\
\\
Provide the scores in a dictionary output. The dictionary is with three keys (name of 3 criterions). The value is a tuple (score and comment).\\
\\
**Example JSON Output:**\\
```json\\
\{\{"Accuracy": (score, "reason"), "Completeness": (score, "reason"), "Reasonableness": (score, "reason")\}\}\\
```\\
\\
**Input Sections:**\\
- **Question \& Answer Pair**: \\
\{qa\}\\
\\
**Output:**\\

\section{Summary of datasets in scientific tasks}
\begin{table*}[!ht]
  \centering
    \caption{Details of datasets and associated tasks they are used to derive instructions, (C for classification; R for regression; T for transformation)}
    \label{table:dataset}
    \begin{tabular}{l|p{8cm}|c}
    \hline
    \textbf{Datasets}   & \textbf{Description}  & \textbf{Task}                 \\ \hline 
      
    \texttt{Glass \cite{Kawazoe1997}} & This dataset is retrieved from a volume of the Landolt– Börnstein collection 'Nonequilibrium phase diagrams of ternary amorphous alloys', containing data on full bulk metallic glass formation ability for 5680 chemical formulas.
    & C \\ \hline

    \texttt{Pei \cite{Pei2020}} & The dataset is from a thermoelectric material review paper published in 2013, which abstracts information from over 100 publications. We used data of 1028 materials from this database. & C \\ \hline

    \texttt{Stability \cite{Batra2020}} &  Consists of water stabilities for over 200 MOFs (metal–organic frameworks), alongside a comprehensive set of chemical features encompassing the metal node, organic ligand, and metal-ligand molar ratios. & C \& T \\ \hline

    \texttt{UV \cite{beard_comparative_2019}} & Auto-generated UV/vis absorption spectra records from 402,034 papers with 18,309 records. We used SMILES to predict category of absorption region (4126 training samples and 1032 test samples). & C \\ \hline

     \texttt{Bandgap \cite{Zhuo2018}} & This dataset is retrieved from Zhuo et al.'s work, containing data on experimental band gaps and DFT calculated zero band gaps for 4604 compounds. & R\\ \hline

     \texttt{Steels \cite{Dunn2020}} & This dataset is retrieved from Citrine informatics, containing data on steel yield strengths from composition for 312 chemical formulas. & R  \\ \hline
    
    \texttt{ChEMBL \cite{Gaulton2012}} & This dataset is sourced from a curated database of bioactive molecules with drug-like properties, focusing on the lipophilicity of 1899 molecular compounds in pharmacokinetics. The water-octanol partition coefficient (logD) is used to describe lipophilicity. & R
    \\ \hline 

    \texttt{Diversity \cite{Moosavi2020} } & A diverse set of structures based on the chemical and geometric descriptors. & R\\ \hline
    
    \texttt{Cp \cite{Moosavi2020}} & Dataset for predicting the heat capacity of materials based on density functional theory simulations. & R \\ \hline

    \texttt{ESOL \cite{Delaney2004}} & This dataset is a compilation of measured aqueous solubility (LogS) values, a crucial factor in drug discovery. The dataset comprises 927 molecular compounds originally used for ESOL - estimated solubility.  & T  \\ \hline 
    
    \end{tabular}
    \end{table*}

\end{document}